\documentclass[sigconf]{acmart}

\usepackage{color}
\usepackage{float}
\usepackage{amsmath}
\usepackage{amssymb}
\usepackage{graphicx}
\usepackage{booktabs}

\floatstyle{ruled}
\newfloat{algorithm}{tbp}{loa}
\providecommand{\algorithmname}{Algorithm}
\floatname{algorithm}{\protect\algorithmname}

\usepackage{listings}

\begin{document}

\title{Multivariate Time Series Classification with WEASEL+MUSE}

\author{Patrick Sch\"afer}
\affiliation{%
	\institution{Humboldt University of Berlin}
	\city{Berlin}
	\state{Germany}
}
\email{patrick.schaefer@informatik.hu-berlin.de}

\author{Ulf Leser}
\affiliation{%
	\institution{Humboldt University of Berlin}
	\city{Berlin}
	\state{Germany}
}
\email{leser@informatik.hu-berlin.de}

\begin{abstract}
Multivariate time series (MTS) arise when multiple interconnected sensors record data over time. 
Dealing with this high-dimensional data is challenging for every classifier for at least two aspects: First, an MTS is not only characterized by individual feature values, but also by the interplay of features in different dimensions. Second, this typically adds large amounts of irrelevant data and noise.
We present our novel MTS classifier WEASEL+MUSE which addresses both challenges. WEASEL+MUSE builds a multivariate feature vector, first using a sliding-window approach applied to each dimension of the MTS, then extracts discrete features per window and dimension. The feature vector is subsequently fed through feature selection, removing non-discriminative features, and analysed by a machine learning classifier. The novelty of WEASEL+MUSE lies in its specific way of extracting and filtering multivariate features from MTS by encoding context information into each feature. Still the resulting feature set is small, yet very discriminative and useful for MTS classification.
Based on a popular benchmark of $20$ MTS datasets, we found that WEASEL+MUSE is among the most accurate classifiers, when compared to the state of the art. The outstanding robustness of WEASEL+MUSE is further confirmed based on motion gesture recognition data, where it out-of-the-box achieved similar accuracies as domain-specific methods.
\end{abstract}

\maketitle
\sloppy

\section{Introduction}
A time series (TS) is a collection of values sequentially ordered in time. TS emerge in many scientific and commercial applications, like weather observations, wind energy forecasting, industry automation, mobility tracking, etc.~\cite{UCRClassification} One driving force behind their rising importance is the sharply increasing use of heterogeneous sensors for automatic and high-resolution monitoring in domains like smart homes~\cite{jerzak2014debs}, machine surveillance~\cite{mutschler2013debs}, or smart grids~\cite{WindPower,hobbs1999analysis}. 

A multivariate time series (MTS) arises when multiple interconnected streams of data are recorded over time. These are typically produced by devices with multiple (heterogeneous) sensors like weather observations (humidity, temperature), Earth movement (3 axis), or satellite images (in different spectra).

In this work we study the problem of multivariate time series classification (MTSC).  Given a concrete MTS, the task of MTSC is to determine which of a set of predefined classes this MTS belongs to, e.g., labeling a sign language gesture based on a set of predefined gestures. The high dimensionality introduced by multiple streams of sensors is very challenging for classifiers, as MTS are not only described by individual features but also by their interplay/co-occurrence in different dimensions~\cite{baydogan2015learning}.

\begin{figure}
	\begin{centering}
		\includegraphics[width=1\columnwidth]{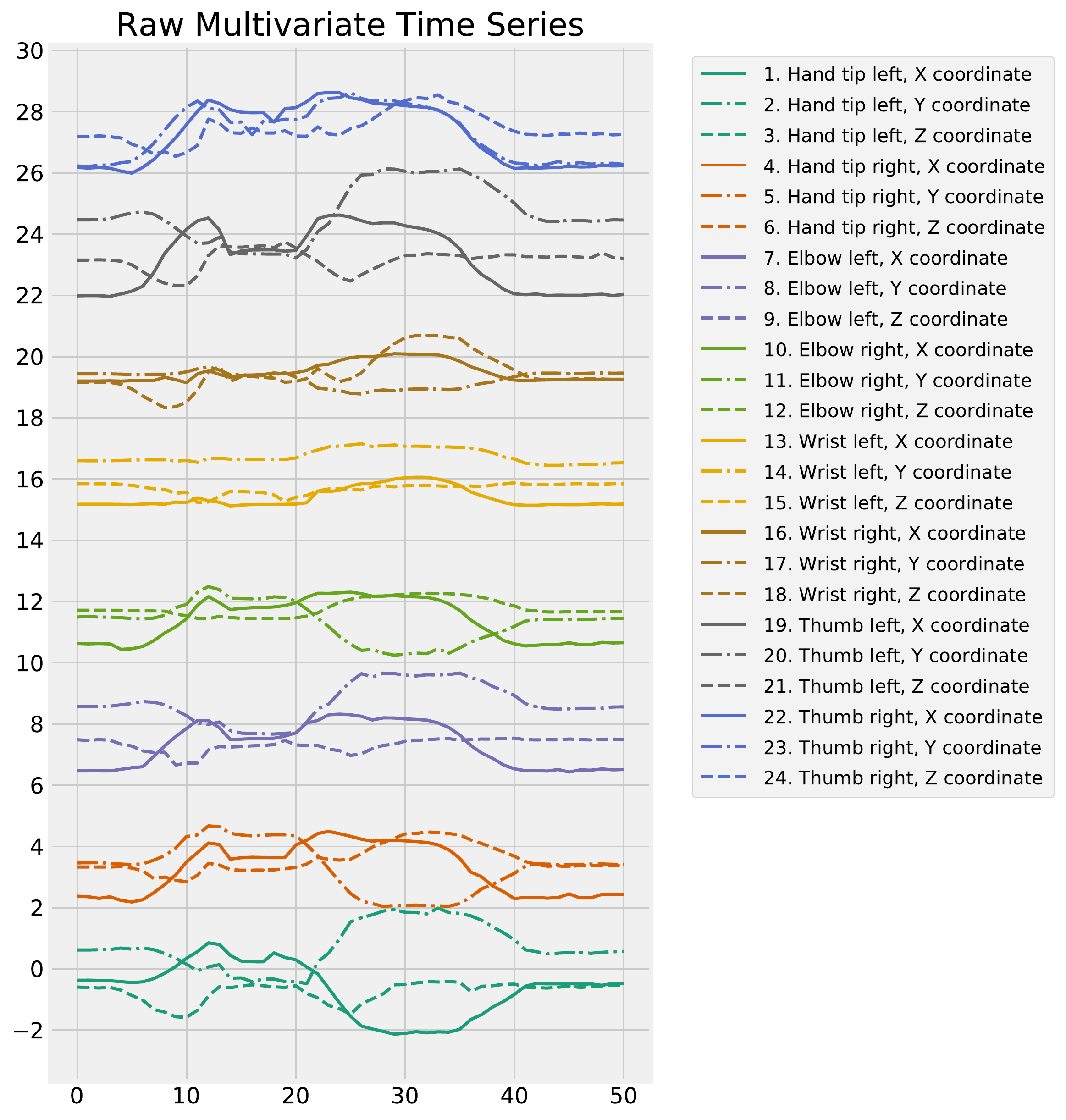}
		\par\end{centering}
	\caption{Motion data recorded from $8$ sensors recording x/y/z coordinates (indicated by different line styles) at the left/right hand, left/right elbow, left/right wrist and left/right thumb (indicated by different colours).\label{fig:model_mts_kinect}}
\end{figure}

As a concrete example, consider the problem of gesture recognition of different users performing isolated gestures (Figure~\ref{fig:model_mts_kinect}). The dataset was recorded using $8$ sensors recording x/y/z coordinates at the left/right hand, left/right elbow, left/right wrist and left/right thumb ($24$ dimensions in total). The data is high dimensional and characterized by long idle periods with small bursts of characteristic movements in every dimension. Here, the exact time instant of an event, e.g., thumbs up, is irrelevant for classification. To effectively deal with this kind of information, an MTSC has to deal with noise, irrelevant dimension data, and, most importantly, extract relevant features from each dimension.

In this paper, we introduce our novel domain agnostic MTSC method called \emph{WEASEL+MUSE (WEASEL plus Multivariate Unsupervised Symbols and dErivatives)}. WEASEL+MUSE conceptually builds on the bag-of-patterns (BOP) model and the \emph{WEASEL (Word ExtrAction for time SEries cLassification)} pipeline for feature selection. The BOP model moves a sliding window over an MTS, extracts discrete features per window, and creates a histogram over feature counts. These histograms are subsequently fed into a machine learning classifier. However, the concrete way of constructing and filtering features in WEASEL+MUSE is different from state-of-the-art multivariate classifiers:

\begin{enumerate}
	\item \textbf{Identifiers}: WEASEL+MUSE adds a dimension (sensor) identifier to each extracted discrete feature. Thereby WEASEL+MUSE can discriminate between the presence of features in different dimensions - i.e., a left vs. right hand was raised. 
		
	\item \textbf{Derivatives}: To improve the accuracy, derivatives are added as features to the MTS. Those are the differences between neighbouring data points in each dimension. 
	These derivatives represent the general shape and are invariant to the exact value at a given time stamp. This information can help to increase classification accuracy.
		
	\item \textbf{Noise robust}: WEASEL+MUSE derives discrete features from windows extracted from each dimension of the MTS using a truncated Fourier transform and discretization, thereby reducing noise.
	
	\item \textbf{Interplay of features}: 
	The interplay of features along the dimensions is learned by 
	assigning weights to features (using logistic regression), thereby boosting or dampening feature counts. Essentially, when two features from different dimensions are characteristic for the class label, these get assigned high weights, and their co-occurrence increases the likelihood of a class.
 
 	\item \textbf{Order invariance}: A main advantage of the BOP model is its invariance to the order of the subsequences, as a result of using histograms over feature counts. Thus, two MTS are similar, if they show a similar number of feature occurrences rather than having the same values at the same time instances.
 	
	\item \textbf{Feature selection}: The wide range of features considered by WEASEL+MUSE (dimensions, derivatives, unigrams, bigrams, and varying window lengths) introduces many non-discriminative features. Therefore, WEASEL+MUSE applies statistical feature selection and feature weighting to identify those features that best discern between classes. The aim of our feature selection is to prune the feature space to a level that feature weighting can be learned in reasonable time.
		
\end{enumerate}

In our experimental evaluation on $20$ public benchmark MTS datasets and a use case on motion capture data, WEASEL+MUSE is constantly among the most accurate methods. 
WEASEL+MUSE clearly outperforms all other classifiers except the very recent deep-learning-based method from \cite{karim2018multivariate}. Compared to the latter, WM performs better for small-sized datasets with less features or samples to use for training, such as sensor readings.


The rest of this paper is organized as follows: Section~2 briefly recaps bag-of-patterns classifiers and definitions. In Section~3 we present related work. In Section~4 we present WEASEL+MUSE's novel way of feature generation and selection. Section~5 presents evaluation results and Section~6 our conclusion.

\section{Background: Time Series and Bag-of-Patterns}

\begin{figure}
	\begin{centering}
		\includegraphics[width=1\columnwidth]{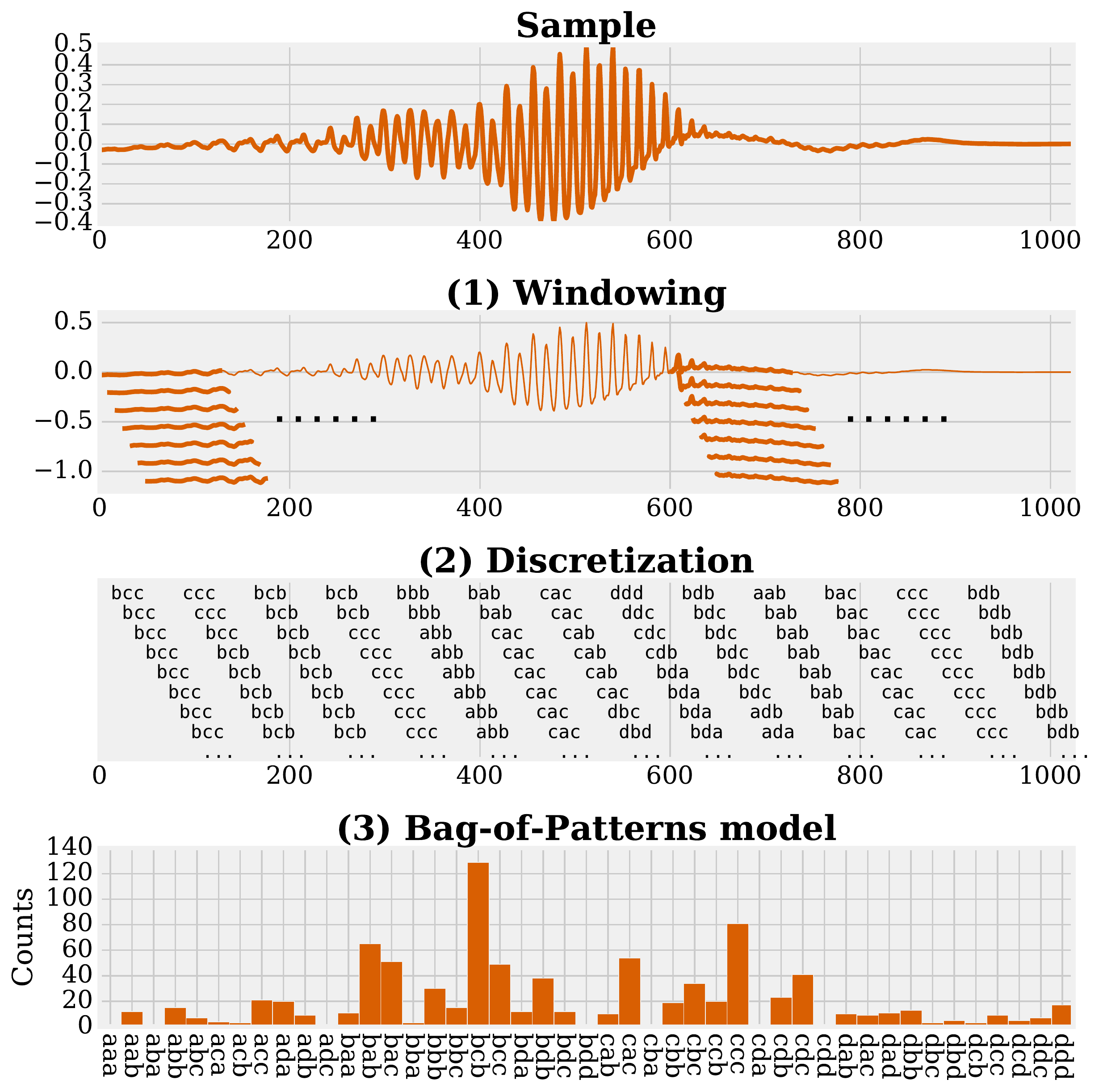}
		\par\end{centering}
	\caption{Transformation of a TS into the Bag-of-Patterns (BOP) model using overlapping windows (second to top), discretization of windows to words (second from bottom), and word counts (bottom) (see~\cite{schaefer2017weasel}).\label{fig:transformation}}
\end{figure}

A \emph{univariate} time series (TS) $T=\{t_{1},\ldots,t_{n}\}$
is an ordered sequence of $n\in\mathbb{N}$ real values $t_{i} \in \mathbb{R}$. A \emph{multivariate} time series (MTS) $T=\{t_{1}, \ldots ,t_{m}\}$ is an ordered sequence of $m \in \mathbb{N}$ streams (dimensions) with $t_{i}=(t_{i,1},\ldots, t_{i,n}) \in \mathbb{R}^{n}$. 
For instance, a stream of $m$ interconnected sensors recording values at each time instant. As we primarily address MTS generated from automatic sensors with a fixed and synchronized sampling along all dimensions, we can safely ignore time stamps. A time series dataset $D$ contains $N$ time series. Note, that we consider only MTS with numerical attributes (not categorical).

The \emph{derivative} of a stream $t_{i}=(t_{i,1},\ldots, t_{i,n})$ is given by the sequence of pairwise differences $t_{i}'=(|t_{i,2}-t_{i,1}|,\ldots, |t_{i,n}-t_{i,n-1}|)$. Adding derivatives to an MTS $T=\{t_{1}, \ldots ,t_{m}\}$ of $m$ streams, effectively doubles the number of streams: $T=\{t_{1}, \ldots ,t_{m}, t'_{1}, \ldots , t'_{m}\}$.

Given a univariate TS $T$, a \emph{window} $S$ of length $w$ is a subsequence with $w$ contiguous values starting at offset $a$ in $T$, i.e., $S(a,w)=(t_{a},\dots,t_{a+w-1})$ with $1\leq a\leq n-w+1$.

We associate each TS with a class label $y\in Y$ from a predefined set of labels $Y$. \emph{Time series classification (TSC)} is the task of predicting a class label for a TS whose label is unknown. A TS classifier is a function that is learned from a set of labelled time series (the training data), that takes an unlabelled time series as input and outputs a label. 

Our method is based on the bag-of-patterns (BOP) model~\cite{0001KL12,schafer2014boss,schafer2015scalable2}. Algorithms following the BOP model build a classification function by (1) extracting subsequences from a TS, (2) discretizing each real valued subsequence into a discrete-valued \emph{word} (a sequence of symbols over a fixed alphabet), (3) building a histogram (feature vector) from word counts, and (4) finally using a classification model from the machine learning repertoire on these feature vectors.

Figure~\ref{fig:transformation} illustrates these steps from a raw time series to a BOP model using overlapping windows. Overlapping subsequences of fixed length are extracted from a time series (second from top), each subsequences is discretized to a word (second from bottom), and finally a histogram is built over the word counts.

Different discretization functions have been used in literature, including SAX~\cite{Lin2007} and SFA~\cite{SchaferH12}. SAX is based on the discretization of mean values and SFA is based on the discretization of coefficients of the Fourier transform. 

In the BOP model, two TS are similar, if the subsequences have similar frequencies in both TS. Feature selection and weighting can be used to damper of emphasize important subsequences, like in the WEASEL model~\cite{schaefer2017weasel}.

\section{Related Work}

Research in univariate TSC has a long tradition and  dozens of approaches have been proposed, refer to~\cite{bagnall2016great,esling2012time,benchmarkingSchaefer17} for summary. The techniques used for TSC can broadly be categorized into two classes: (a) similarity-based (distance-based) methods and (b) feature-based methods. 

\emph{Similarity-based} methods make use of a similarity measure like Dynamic Time Warping (DTW)~\cite{rakthanmanon2012searching} to compare two TS. 1-Nearest Neighbour DTW is commonly used as a baseline in TSC comparisons~\cite{bagnall2016great}. In contrast, \emph{feature-based} TSC rely on comparing features, typically generated from substructures of a TS. The most successful approaches are \emph{shapelets} or \emph{bag-of-patterns} (BOP). \emph{Shapelets} are defined as TS subsequences that are maximally representative of a class~\cite{YeK09}. The standard BOP model~\cite{0001KL12} breaks up a TS into windows, represent these as discrete features, and finally build a histogram of feature counts as basis for classification. 

In previous research we have studied the BOP model for univariate TSC. The BOSS (Bag-of-SFA-Symbols)~\cite{schafer2014boss} classifier is based on the (unsupervised) Symbolic Fourier Approximation (SFA)~\cite{SchaferH12} to generate discrete features and uses a similarity measure on the histogram of feature counts. The WEASEL classifier~\cite{schaefer2017weasel} applies a supervised symbolic representation to transform subsequences to words, uses statistical feature selection, and subsequently feeds the words into a logistic regression classifier. WEASEL is among the most accurate and fastest univariate TSC~\cite{schaefer2017weasel}. WEASEL was optimized to extract discriminative words to ease classification of univariate TS. We observed that this led to an overall low accuracy for MTSC due to the increased number of possible features along all dimensions (see Section~5). WEASEL+MUSE was designed on the WEASEL pipeline, but adding sensor identifiers to each word, generating unsupervised discrete features to minimize overfitting, as opposed to WEASEL that uses a supervised transformation. WEASEL+MUSE further adds derivatives (differences between all neighbouring points) to the feature space to increase accuracy.
 
For multivariate time series classification (MTSC), the most basic approach is to apply rigid dimensionality reduction (i.e., PCA) or simply concatenate all dimensions of the MTS to obtain a univariate TS and use proven univariate TSC. Some domain agnostic MTSC have been proposed. 

\emph{Symbolic Representation for Multivariate Time series (SMTS)}~\cite{baydogan2015learning} uses codebook learning and the bag-of-words (BOW) model for classification. First, a random forest is trained on the raw MTS to partition the MTS into leaf nodes. Each leaf node is then labelled by a word of a codebook. There is no additional feature extraction, apart from calculating  derivatives for the numerical dimensions (first order differences). For classification a second random forest is trained on the BOW representation of all MTS. 

\emph{Ultra Fast Shapelets (UFS)}~\cite{wistuba2015ultra} applies the shapelet discovery method to MTS classification. The major limiting factor for shapelet discovery is the time to find discriminative subsequences, which becomes even more demanding when dealing with MTS. UFS solves this by extracting random shapelets. On this transformed data, a linear SVM or a Random Forest is trained. Unfortunately, the code is not available to allow for reproducibility 

\emph{Generalized Random Shapelet Forests (gRSF)}~\cite{karlsson2016generalized} also generates a set of shapelet-based decision trees over randomly extracted shapelets. In their experimental evaluation, gRSF was the best MTSC when compared to SMTS, LPS and UFS on 14 MTS datasets. Thus, we use gRFS as a representative for random shapelets.

\emph{Learned Pattern Similarity (LPS)}~\cite{baydogan2016time} extracts segments from an MTS. It then trains regression trees to identify structural dependencies between segments. The regression trees trained in this	manner represent a non-linear AR model. LPS next builds a BOW representation based on the labels of the leaf nodes similar to SMTS. Finally a similarity measure is defined on the BOW representations of the MTS. LPS showed better performance than DTW in a benchmark using 15 MTS datasets.
\emph{Autoregressive (AR) Kernel}~\cite{cuturi2011autoregressive} proposes an AR kernel-based distance measure for MTSC.

\emph{Autoregressive forests for multivariate time series modelling (mv-ARF)}~\cite{tuncel2018autoregressive} proposes a tree ensemble trained on autoregressive models, each one with a different lag, of the MTS. 
This model is used to capture linear and non-linear relationships between features in the dimensions of an MTS.
The authors compared mv-ARF to AR Kernel, LPS and DTW on $19$ MTS datasets. mv-ARF and AR kernel showed the best results. mv-ARF performs well on motion recognition data. AR kernel outperformed the other methods for sensor readings.

At the time of writing this paper, \emph{Multivariate LSTM-FCN}~\cite{karim2018multivariate} was proposed that introduces a deep learning architecture based on a long short-term memory  (LSTM), a fully convolutional network (FCN) and a squeeze and excitation block. Their method is compared to state-of-the-art and shows the overall best results. 



\section{WEASEL+MUSE}\label{sec:TSCMUSE}

\begin{figure*}
	\includegraphics[width=2\columnwidth]{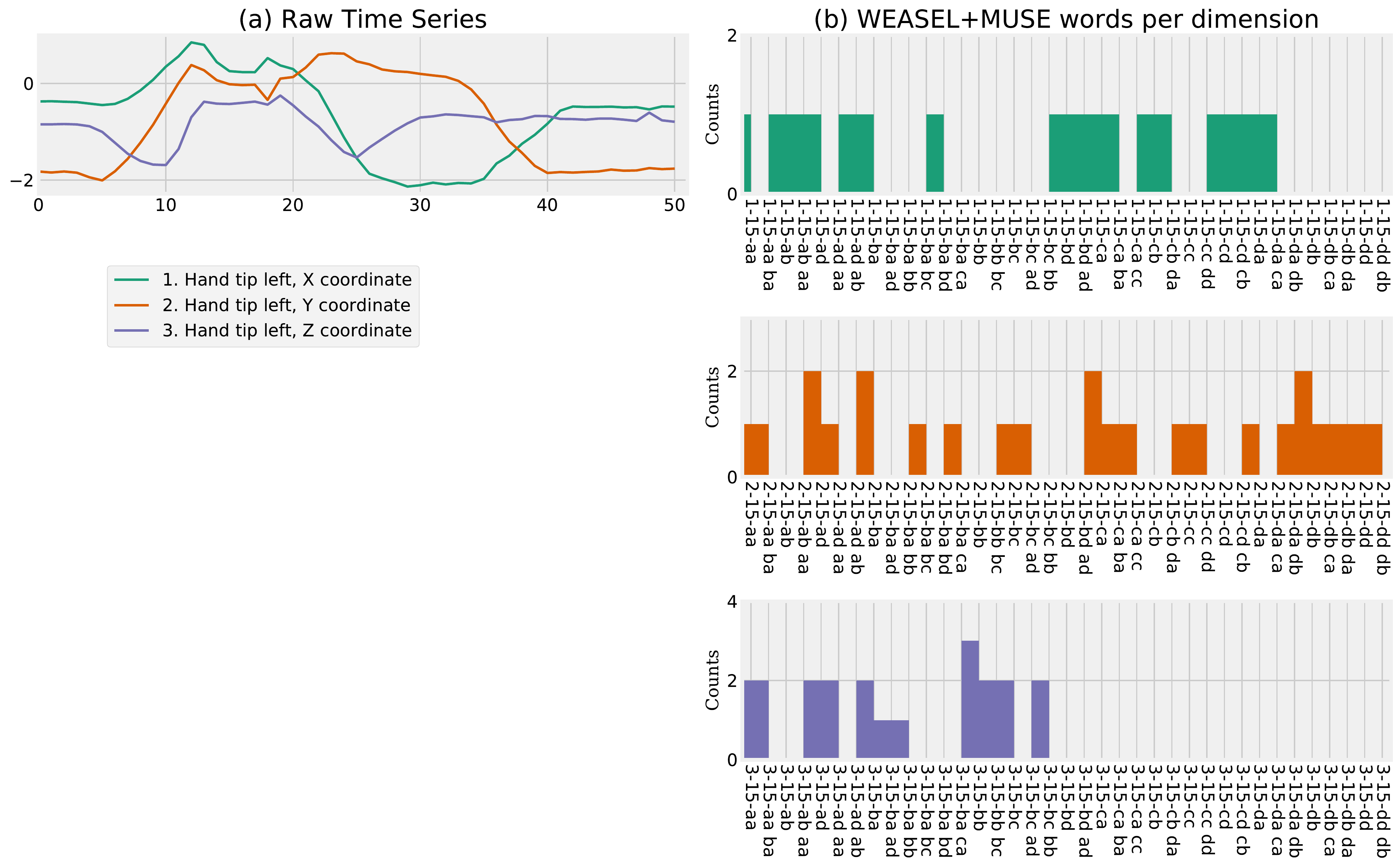}
	\caption{WEASEL+MUSE model of a motion capture. (a) motion of a left hand in x/y/z coordinates. (b) the WEASEL+MUSE model for each of these coordinates. A feature in the WEASEL+MUSE model encodes the dimension, window length and actual word, e.g., $1\_15\_aa$ for 'left Hand', window length $15$ and word 'aa'.\label{fig:MUSE-Model}}
\end{figure*}

We present our novel method for domain agnostic multivariate time series classification (MTSC) called \emph{WEASEL+MUSE (WEASEL+Multivariate Unsupervised Symbols and dErivatives)}. WEASEL+MUSE addresses the major challenges of MTSC in a specific manner (using gesture recognition as an example):

\begin{enumerate}
	\item \emph{Interplay of dimensions}: MTS are not only characterized by individual features at a single time instance, but also by the interplay of features in different dimensions. For example, to predict a hand gesture, a complex orchestration of interactions between hand, finger and elbow may have to be considered. 
	
	\item \emph{Phase invariance}: Relevant events in an MTS do not necessarily reappear at the same time instances in each dimension. Thus, characteristic features may appear anywhere in an MTS (or not at all). For example, a hand gesture should allow for considerable differences in time schedule.
		
	\item \emph{Invariance to irrelevant dimensions}: Only small periods in time and in some streams may contain relevant information for classification. What makes things even harder is the fact that whole sensor streams may be irrelevant for classification. For instance, a movement of a leg is irrelevant to capture hand gestures and vice versa. 
\end{enumerate}

We engineered WEASEL+MUSE to address these challenges. Our method conceptually builds on our previous work on the bag-of-patterns (BOP) model and univariate TSC~\cite{schafer2014boss,schaefer2017weasel}, yet uses a different approach in many of the individual steps to deal with the aforementioned challenges. We will use the terms \emph{feature} and \emph{word} interchangeably throughout the text. 
In essence, WEASEL+MUSE makes use of a histogram of feature counts. In this feature vector it captures information about local and global changes in the MTS along different dimensions. It then learns weights to boost or damper characteristic features. The interplay of features is represented by high weights.

\subsection{Overview}

We first give an overview of our basic idea and an example how we deal with the challenges described above. In WEASEL+MUSE a feature is represented by a word that encodes the identifiers (sensor id, window size, and discretized Fourier coefficients) and counts its occurrences. Figure~\ref{fig:MUSE-Model}  shows an example for the WEASEL+MUSE model of a fixed window length $15$ on motion capture data. The data has 3 dimensions (x,y,z coordinates). A feature $('3\_15\_ad', 2)$ (see Figure~\ref{fig:MUSE-Model} (b)) represents a unigram 'ad' for the z-dimension with window length $15$ and frequency $2$, or the feature $('2\_15\_bd\_ad', 2)$ represents a bigram 'bd\_ad' for the y-dimension with length $15$ and frequency $2$.

\paragraph{Pipeline:}

\begin{figure*}
	\includegraphics[width=2\columnwidth]{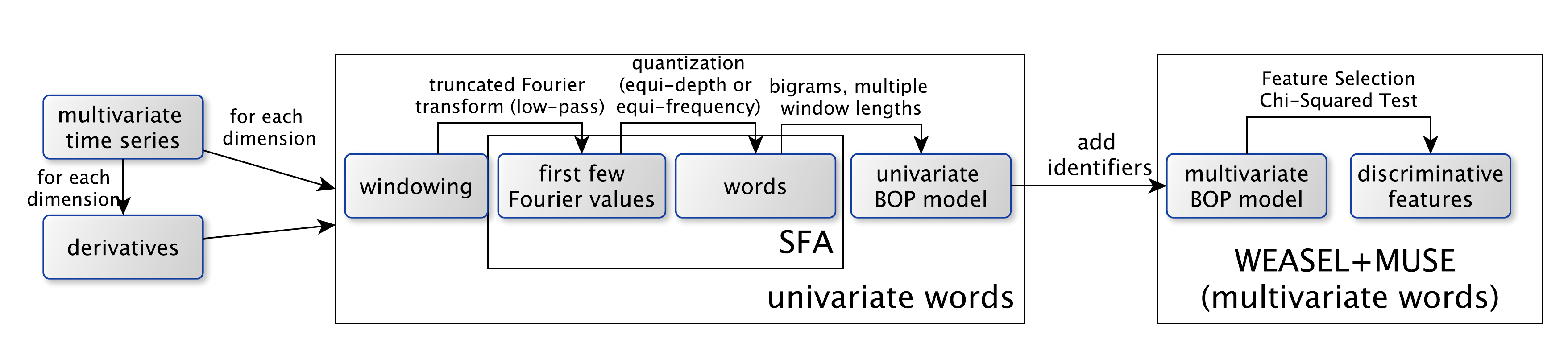}
	\caption{WEASEL+MUSE Pipeline: Feature extraction, univariate Bag-of-Patterns (BOP) models and WEASEL+MUSE.\label{fig:MUSE-Pipeline}}
\end{figure*}

WEASEL+MUSE is composed of the building blocks depicted in Figure~\ref{fig:MUSE-Pipeline}: the symbolic representation SFA~\cite{SchaferH12}, BOP models for each dimension, feature selection and the WEASEL+MUSE model. 
WEASEL+MUSE conceptionally builds upon the univariate BOP model applied to each dimension. Multivariate words are obtained from the univariate words of each BOP model by concatenating each word with an identifier (representing the sensor and the window size). This maintains the association between the dimension and the feature space. 

More precisely, an MTS is first split into its dimensions. Each dimension can now be considered as a univariate TS and transformed using the classical BOP approach. To this end, z-normalized windows of varying lengths are extracted. Next, each window is approximated using the truncated Fourier transform, keeping only lower frequency components of each window. Fourier values (real and imaginary part separately) are then discretized into words based on equi-depth or equi-frequency binning using a symbolic transformation (details will be given in Subsection~\ref{subsec:SymbolicRepresentation}). Thereby, words (unigrams) and pairs of words (bigrams) with varying window lengths are computed. These words are concatenated with their identifiers, i.e., the sensor id (dimension) and the window length.
Thus, WEASEL+MUSE keeps a disjoint word space for each dimension and two words from different dimensions can never coincide. To deal with irrelevant features and dimensions, a Chi-squared test is applied to all multivariate words (Subsection~\ref{subsec:Chi-Squared-Test}). As a result, a highly discriminative feature vector is obtained and a fast linear time logistic regression classifier can be trained (Subsection~\ref{subsec:Chi-Squared-Test}). It further captures the interplay of features in different dimensions by learning high weights for important features in each dimension (Subsection~\ref{subsec:Feature-interplay}).

\subsection{Word Extraction: Symbolic Fourier Approximation}\label{subsec:SymbolicRepresentation}

\begin{figure}
	\begin{centering}
		\includegraphics[width=1\columnwidth]{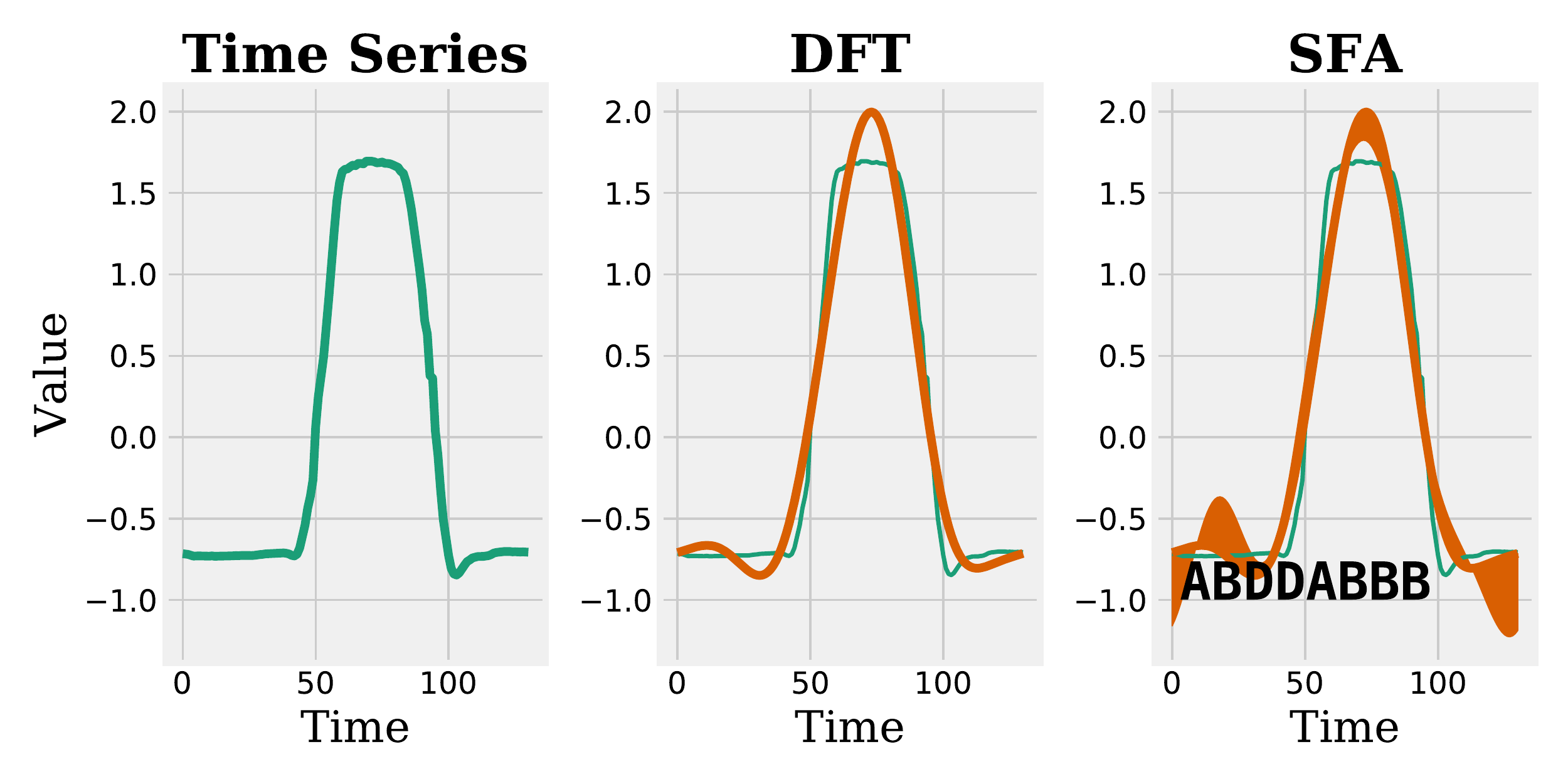}
	\end{centering}
	\caption{The Symbolic Fourier Approximation (SFA): A time series~(left) is approximated using the truncated Fourier transform~(centre) and discretized to the word \emph{ABDDABBB}~(right) with the four-letter alphabet ('a' to 'd'). The inverse transform is depicted by an orange area (right), representing the tolerance for all signals that will be mapped to the same word.\label{fig:SFATransform}}
\end{figure}

Instead of training a multivariate symbolic transformation, we train and apply the univariate symbolic transformation SFA to each dimension of the MTS separately. This allows for (a) phase invariance between different dimensions, as a separate BOP model is built for each dimension, but (b) the information that two features occurred at \emph{exactly} the same time instant in two different dimensions is lost.
Semantically, splitting an MTS into its dimensions results in two MTS $T_1$ and $T_2$ to be similar, if both share similar substructures within the $i$-th dimension at \emph{arbitrary} time stamps. 

SFA transforms a real-valued TS window to a word using an alphabet of size $c$ as in~\cite{SchaferH12}: 
 
 \begin{enumerate}
	\item \emph{Approximation:} Each normalized window of length $w$ is subjected to dimensionality reduction by the use of the truncated Fourier transform, keeping only the first $l \ll w$ coefficients for further analysis. This step acts as a low pass (noise) filter, as higher order Fourier coefficients typically represent rapid changes like drop-outs or noise.
 	\item \emph{Quantization:} Each Fourier coefficient is then discretized to a symbol of an alphabet of fixed size $c$, which in turn achieves further robustness against noise.
 \end{enumerate}
 
Figure~\ref{fig:SFATransform} exemplifies this process for a univariate time series, resulting in the word \emph{ABDDABBB}. 
As a result, each real-valued window in the $i$-th dimension is transformed into a word of length $l$ with an alphabet of size $c$. For a given window length, there are a maximum of $\mathcal{O}(n)$ windows in each of the $m$ dimensions, resulting in a total of $\mathcal{O}(n \times m)$ words.
  
SFA is a data-adaptive symbolic transformation, as opposed to SAX~\cite{Lin2007} which always uses the same set of bins irrelevant of the data distribution. Quantization boundaries are derived from a (sampled) train dataset using either (a) \emph{equi-depth} or (b) \emph{equi-frequency} binning, such that (a) the Fourier frequency range is divided into equal-sized bins or (b) the boundaries are chosen to hold an equal number of Fourier values. SFA is trained for each dimension separately, resulting in $m$ SFA transformations. Each SFA transformation is then used to transform only its dimension of the MTS.


\subsection{Univariate Bag-of-Patterns: Unigrams, bigrams, derivatives, window lengths}\label{subsec:BOP}

In the BOP model, two TS are distinguished by the frequencies of certain subsequences rather than their presence or absence. A TS is represented by word counts, obtained from the windows of the time series. BOP-based methods have a number of parameters, and of particular importance is the window length, which heavily influences its performance. For dealing with MTS, we have to find the best window lengths for each dimension, as one cannot assume that there is a single optimal value for all dimensions. 
WEASEL+MUSE addresses this issue by building a large feature space using multiple window lengths, the MTS dimensions, unigrams, bigrams, and derivatives. This very large feature space is aggressively reduced in a second separate step~\ref{subsec:Chi-Squared-Test}. 

The feature set of WEASEL+MUSE, given an MTS $T=(t_1,\dots,t_m)$ is composed of (see also Section~\ref{subsec:Chi-Squared-Test}):
\begin{enumerate}
	\item \emph{Derivatives:} Derivatives are added to the MTS. These are the differences between all neighbouring points in one dimension (see Section~2). This captures information about how much a signal changes in time. It has been shown that this additional information can improve the accuracy~\cite{baydogan2015learning}. We show the utility of derivatives in Section~\ref{subsec:influence}.
			
	\item \emph{Local and Global Substructures:} For each possible window lengths $w \in [4..len(t_i)]$, windows are extracted from the dimensions and the derivatives, and each window is transformed to a word using the SFA transformation. This helps to capture both local and global patterns in an MTS. 
	
	\item \emph{Unigrams and Bigrams:} Once we have extracted all words (unigrams), we enrich this feature space with co-occurrences of words (bigrams). It has been shown in ~\cite{schaefer2017weasel} that the usage of bigrams reduces the order-invariance of the BOP model. We could include m-grams, but the feature space grow polynomial in the m-gram number, such that it is infeasible to use anything larger than bigrams (resulting in $O(n^2)$ features).
	
	\item \emph{Identifiers:} Each word is concatenated with it's sensor id and window size (see Figure~\ref{fig:MUSE-Model}). It is rather meaningless to compare features from different sensors: if a temperature sensor measures 10 and a humidity sensor measures 10, these capture totally different concepts. To distinguish between sensors, the features are appended with sensor ids. e.g., (temp: 10) and (humid: 10). However, both measurements can be important for classification. Thus, we add them to the same feature vector and use feature selection and feature weights to identify the important ones.
\end{enumerate}

\paragraph{Pseudocode:}

\begin{algorithm}[t]
	\begin{lstlisting}[language=Java,numbers=left,basicstyle={\sffamily},breaklines=true,showstringspaces=false,tabsize=2,numbersep=1em,xleftmargin=2em,xrightmargin=0em,emph={function, in, all, each, to},emphstyle={\textbf},escapechar={ß}]
function ß\textbf{WEASEL\_MUSE}ß(mts, l, wLen)
	bag = empty BagOfPattern
	// extract from each dimension
	for each dimId in mts:
	  // use multiple window lenghts
	  for each window-length w in wLen:
	    for each (prevWindow, window) in ß\texttt{SLIDING\_WINDOWS}ß(mts[dimId], wLen):
	
	    // BOP computed from unigrams
	    word = ß\texttt{SFA}ß(window,l)
	    unigram = concat(dim,w,word)
    	bag[unigram].increaseCounts()
	
	    // BOP computed from bigrams
    	prevWord=ß\texttt{SFA}ß(prevWindow,l)	
    	bigram =concat(dim,w,prevWord,word)
    	bag[bigram].increaseCounts()
    	
	// feature selection using ChiSquared
	return ß\texttt{CHI\_SQUARED\_FILTER}ß(bag)
	\end{lstlisting}
	
	\caption{Build one BOP model using SFA,
		multiple window lengths, bigrams and the Chi-squared test for feature
		selection. $l$ is the number of Fourier values to keep and $wLen$ are the window lengths used for sliding window extraction.\label{alg:The-WEASEL-representation}}
\end{algorithm}

Algorithm~\ref{alg:The-WEASEL-representation} illustrates WEASEL+MUSE: sliding windows of length $w$ are extracted in each dimension (line~7). We empirically set the window lengths to all values in $[4,\dots,n]$. Smaller values are possible, but the feature space can become untraceable, and small window lengths are basically meaningless for TS of length $>10^3$.
The SFA transformation is applied to each real-valued sliding window (line~10,15). Each word is concatenated with the window length and dimension identifier, and its occurrence is counted (line~12,17). Lines~15\textendash17 illustrate the use of bigrams: the preceding sliding window is concatenated with the current window. Note, that all words (each dimension, unigrams, bigrams, each window-length) are joined within a single bag-of-patterns. 
Finally irrelevant words are removed from this bag-of-patterns using the Chi-squared test~(line~20). The target SFA length $l$ is learned through cross-validation.

\subsection{Feature Selection and Weighing: Chi-squared Test and Logistic Regression}\label{subsec:Chi-Squared-Test}

WEASEL+MUSE applies the Chi-squared ($\chi^{2}$) test to identify the most relevant features, only features passing a certain threshold are kept to reduce this feature space prior to training the classifier. We set the threshold such that it is high enough for the logistic regression classifier to train a model in reasonable time (and when set too low, training takes longer). If a feature is irrelevant but not removed due to the $\chi^{2}$-test, it will get assigned a low weight by the logistic regression classifier. It would be possible to use different feature selection methods. As our  main aim is to reduce the runtime for training, we did not look into other feature selection techniques.

For a set of $N$ $m$-dimensional MTS of length $n$, the size of the BOP feature space is $\mathcal{O}(\min(Nn^2, c^l)\times m)$ for word length $l$, $c$  symbols and $m$ dimensions. The number of MTS $N$ and length $n$ affects the actual word frequencies. But in the worst case each TS window can only produce one distinct word, and there are $Nn^2$ windows in each dimension. WEASEL+MUSE further uses bigrams, derivatives, and $\mathcal{O}(n)$ window lengths. WEASEL+MUSE keeps a disjoint word space for each dimension and window lengths, thus two words from different dimensions can never collide (no false positives). Thus, the theoretical dimensionality of this feature space rises to $\mathcal{O}(\min [Nn^2,c^{2l}\cdot~n] \times m)$. Essentially, the feature space can grow quadratically with the number of observations of an MTS, if every observation generates a distinct word. However, in practice we never observed that many features due to the periodicity of TS or superfluous data/dimensions. Statistical feature selection reduces the total number of features to just a few hundred features.

We use sparse vectors to store the words for each MTS, as each feature vector only contains a few features after feature selection. We implemented our MTS classifier using liblinear~\cite{fan2008liblinear} as it scales linearly with the dimensionality of the feature space~\cite{ng2004feature}.

\subsection{Feature Interplay}\label{subsec:Feature-interplay}

The WEASEL+MUSE model is essentially a histogram of discrete features extracted from all dimensions. The logistic regression classifier trains for each class a weight vector, to assign high weights to those features that are relevant within each dimension. Thereby, it captures the feature interplay, as dimensions are not treated separately but the weight vector is trained over all dimensions. Still, this approach allows for phase-invariance as classes (events) are represented by the frequency of occurrence of discrete features rather than the exact time instance of an event.

\section{Evaluation}\label{sec:Results}
\subsection{Experimental Setup}

\begin{table}[t]
	\begin{centering}
		\begin{tabular}{cccccc}
			& {\#classes} & {m} & {n} & {N Train} & {N Test}\tabularnewline
			\midrule 
			\midrule 
			{Digits} & {10} & {13} & {4-93} & {6600} & {2200}\tabularnewline
			\midrule 
			{AUSLAN} & {95} & {22} & {45-136} & {1140} & {1425}\tabularnewline
			\midrule 
			{CharTrajectories} & {20} & {3} & {109-205} & {300} & {2558}\tabularnewline
			\midrule 
			{CMUsubject16} & {2} & {62} & {127-580} & {29} & {29}\tabularnewline
			\midrule 
			{DigitShapes} & 4 & {2} & {30-98} & {24} & {16}\tabularnewline
			\midrule 
			{ECG} & {2} & {2} & {39-152} & {100} & {100}\tabularnewline
			\midrule 
			{Japanese Vowels} & {9} & {12} & {7-29} & {270} & {370}\tabularnewline
			\midrule 
			{KickvsPunch} & {2} & {62} & {274-841} & {16} & {10}\tabularnewline
			\midrule 
			{LIBRAS} & {15} & {2} & {45} & {180} & {180}\tabularnewline
			\midrule 
			{Robot Failure LP1} & {4} & {6} & {15} & {38} & {50}\tabularnewline
			\midrule 
			{Robot Failure LP2} & {5} & {6} & {15} & {17} & {30}\tabularnewline
			\midrule 
			{Robot Failure LP3} & {4} & {6} & {15} & {17} & {30}\tabularnewline
			\midrule 
			{Robot Failure LP4} & {3} & {6} & {15} & {42} & {75}\tabularnewline
			\midrule 
			{Robot Failure LP5} & {5} & {6} & {15} & {64} & {100}\tabularnewline
			\midrule 
			{NetFlow} & {2} & {4} & {50-997} & {803} & {534}\tabularnewline			
			\midrule 
			{PenDigits} & {10} & {2} & {8} & {300} & {10692}\tabularnewline
			\midrule 
			{Shapes} & {3} & {2} & {52-98} & {18} & {12}\tabularnewline
			\midrule 			
			{UWave} & {8} & {3} & {315} & {200} & {4278}\tabularnewline
			\midrule 
			{Wafer} & {2} & {6} & {104-198} & {298} & {896}\tabularnewline
			\midrule 
			{WalkvsRun} & {2} & {62} & {128-1918} & {28} & {16}\tabularnewline
		\end{tabular}
		\par\end{centering}
	\caption{$20$ multivariate time series datasets collected from~\cite{MultivariateTimeSeriesClassification}. \label{tab:Datasets}}
	
\end{table}

\begin{table*}[t]
	\begin{centering}
		\begin{tabular*}{2\columnwidth}{@{\extracolsep{\fill}}cccccccccc}
						Dataset & SMTS & LPS & mvARF & DTWi & ARKernel & gRSF & MLSTMFCN & MUSE\tabularnewline
			\hline
			\hline 
			ArabicDigits & 96.4\% & 97.1\% & 95.2\% & 90.8\% & 98.8\% & 97.5\% & 99.0\% & \textbf{99.2\%}\tabularnewline
			\hline 
			AUSLAN & 94.7\% & 75.4\% & 93.4\% & 72.7\% & 91.8\% & 95.5\% & 95.0\% & \textbf{97\%}\tabularnewline
			\hline 
			CharTrajectories & 99.2\% & 96.5\% & 92.8\% & 94.8\% & 90\% & 99.4\% & 99.0\% & 97.3\%\tabularnewline
			\hline 
			CMUsubject16 & 99.7\% & \textbf{100\%} & \textbf{100\%} & 93\% & \textbf{100\%} & \textbf{100\%} & \textbf{100\%} & \textbf{100\%}\tabularnewline
			\hline 
			ECG & 81.8\% & 82\% & 78.5\% & 79\% & 82\% & \textbf{88\%} & 87\% & \textbf{88\%}\tabularnewline
			\hline 
			JapaneseVowels & 96.9\% & 95.1\% & 95.9\% & 96.2\% & 98.4\% & 80\% & \textbf{100\%} & 97.6\%\tabularnewline
			\hline 
			KickvsPunch & 82\% & 90\% & 97.6\% & 60\% & 92.7\% & \textbf{100\%} & 90\% & \textbf{100\%}\tabularnewline
			\hline 
			Libras & 90.9\% & 90.3\% & 94.5\% & 88.8\% & 95.2\% & 91.1\% & \textbf{97\%} & 89.4\%\tabularnewline
			\hline 
			NetFlow & \textbf{97.7\%} & 96.8\% & NaN & 97.6\% & NaN & 91.4\% & 95\% & 96.1\%\tabularnewline
			\hline 
			UWave & 94.1\% & \textbf{98\%} & 95.2\% & 91.6\% & 90.4\% & 92.9\% & 97\% & 91.6\%\tabularnewline
			\hline 
			Wafer & 96.5\% & 96.2\% & 93.1\% & 97.4\% & 96.8\% & 99.2\% & 99\% & \textbf{99.7\%}\tabularnewline
			\hline 
			WalkvsRun & \textbf{100\%} & \textbf{100\%} & \textbf{100\%} & \textbf{100\%} & \textbf{100\%} & \textbf{100\%} & \textbf{100\%} & \textbf{100\%}\tabularnewline
			\hline 
			LP1 & 85.6\% & 86.2\% & 82.4\% & 76\% & 86\% & 84\% & 80\% & \textbf{94\%}\tabularnewline
			\hline 
			LP2 & 76\% & 70.4\% & 62.6\% & 70\% & 63.4\% & 66.7\% & \textbf{80\%} & 73.3\%\tabularnewline
			\hline 
			LP3 & 76\% & 72\% & 77\% & 56.7\% & 56.7\% & 63.3\% & 73\% & \textbf{90\%}\tabularnewline
			\hline 
			LP4 & 89.5\% & 91\% & 90.6\% & 86.7\% & 96\% & 86.7\% & 89\% & \textbf{96\%}\tabularnewline
			\hline 
			LP5 & 65\% & \textbf{69\%} & 68\% & 54\% & 47\% & 45\% & 65\% & \textbf{69\%}\tabularnewline
			\hline 
			PenDigits & 91.7\% & 90.8\% & 92.3\% & 92.7\% & 95.2\% & 93.2\% & \textbf{97\%} & 91.2\%\tabularnewline
			\hline 
			Shapes & \textbf{100\%} & \textbf{100\%} & \textbf{100\%} & \textbf{100\%} & \textbf{100\%} & \textbf{100\%} & \textbf{100\%} & \textbf{100\%}\tabularnewline
			\hline 
			DigitShapes & \textbf{100\%} & \textbf{100\%} & \textbf{100\%} & \textbf{93.8\%} & \textbf{100\%} & \textbf{100\%} & \textbf{100\%} & \textbf{100\%}\tabularnewline
			\hline
			\hline 
			Wins/Ties & 4 & 6 & 4 & 2 & 5 & 6 & 8 & \textbf{13}\tabularnewline
			\hline 
			Mean & 90.7\% & 89.8\% & 90\% & 84.6\% & 88.4\% & 88.7\% & 92.1\% & \textbf{93.5\%}\tabularnewline
			\hline 
			Avg. Rank & $4.05$ & $4.05$ & $4.7$ & $6.6$ & $4.35$ & $3.85$ & $2.75$ & $2.45$ \tabularnewline 
		\end{tabular*}
		\par\end{centering}	
	\caption{Accuracies for each dataset. The best approaches are highlighted using a bold font.\label{tab:accuracies}}
\end{table*}

\begin{itemize}
	\item \emph{Datasets:} We evaluated our \emph{WEASEL+MUSE} classifier using 20 publicly available MTS dataset listed in Table~\ref{tab:Datasets}. Furthermore, we compared its performance on a real-life dataset taken from the motion capture domain; results are reported in Section~\ref{subsec:usecase}. Each MTS dataset provides a train and test split set which we use unchanged to make our results comparable to prior publications. 
	\item \emph{Competitors:} We compare WEASEL+MUSE to the $7$ domain agnostic state-of-the-art MTSC methods we are aware of ARKernel~\cite{cuturi2011autoregressive}, LPS~\cite{baydogan2016time}, mv-ARF~\cite{tuncel2018autoregressive}, SMTS~\cite{baydogan2015learning}, gRSF~\cite{karlsson2016generalized}, MLSTM-FCN~\cite{karim2018multivariate}, and the common baseline Dynamic Time Warping independent (DTWi), implemented as the sum of DTW distances in each dimension with a full warping window. We use the reported test accuracies on the MTS datasets given by the authors in their publications, thereby avoiding any bias in training settings parameters. All reported numbers in our experiments correspond to the accuracy on the test split. 
	We were not able to reproduce the published results for MLSTM-FCN using their code. The authors told us that this is due to random seeding and their results are based on a single run. Instead, we report the median over $5$ runs using their published code~\cite{karim2018multivariate}.	
	For SMTS and gRSF, we additionally ran their code on the missing $5$ and $7$ datasets. The code for UFS is not available, thus we did not include it into the experiments.
	The web-page\footnote{http://www.timeseriesclassification.com} lists state-of-the-art univariate TSC. However, we cannot use univariate TSC to classify multivariate datasets.
	\item \emph{Server:} The experiments were carried out on a server running LINUX with 2xIntel Xeon E5-2630v3 and 64GB RAM, using JAVA JDK x64 1.8. 
	\item \emph{Training WEASEL+MUSE:} For WEASEL+MUSE we performed $10$-fold cross-validation on the train datasets to find the most appropriate parameters for the SFA word lengths $l\in\left[2,4,6\right]$ and SFA quantization method \emph{equi-depth} or \emph{equi-frequency} binning.  All other parameters are constant: $chi=2$, as we observed that varying these values has only a negligible effect on the accuracy. We used \emph{liblinear} with default parameters $(\textit{bias}=1, p=0.1, c=5$ and solver L2R\_LR\_DUAL). To ensure reproducible results, we provide the WEASEL+MUSE source code and the raw measurement sheets~\cite{MUSEWebPage}. 
\end{itemize}

\subsection{Accuracy}

\begin{figure}[t]
	\begin{centering}
		\includegraphics[width=1.0\columnwidth]{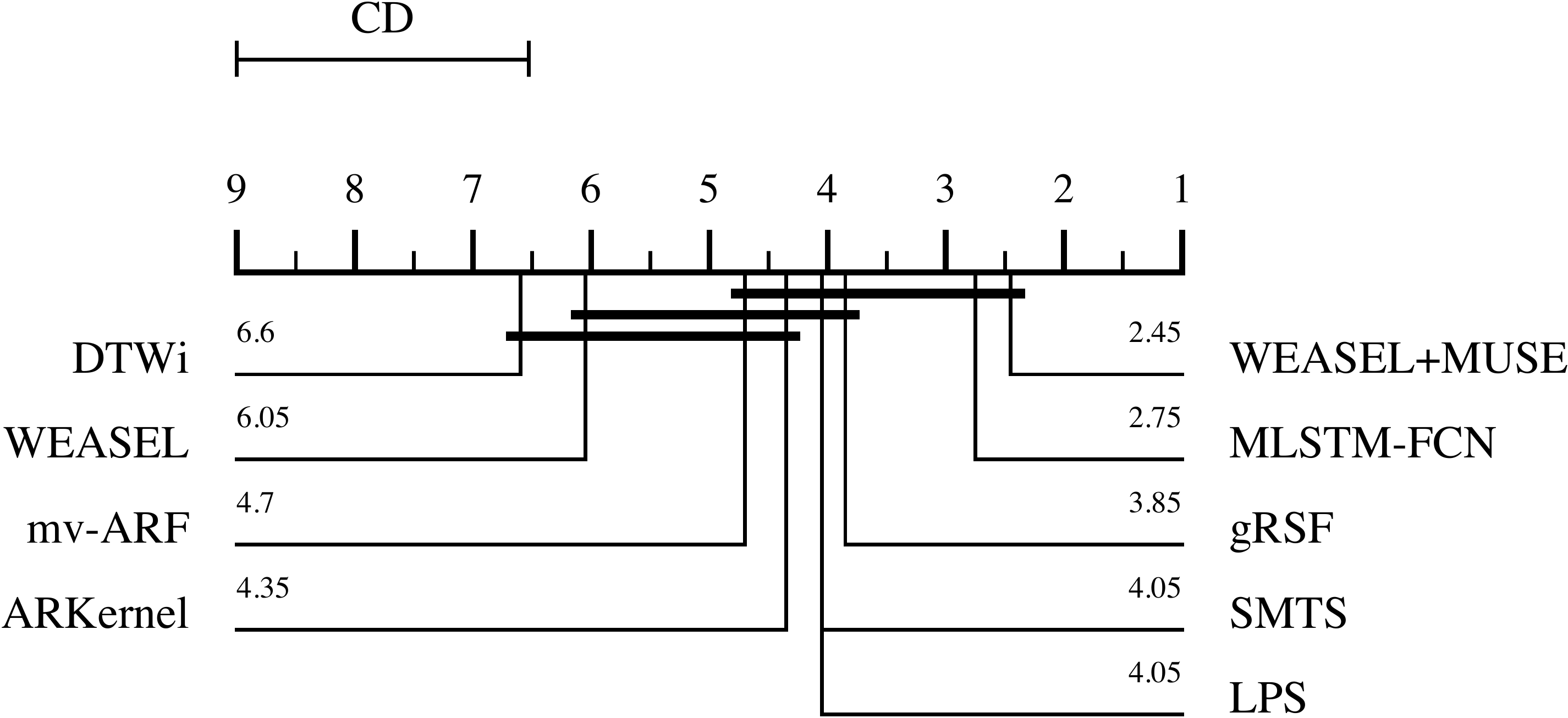}
	\end{centering}
	\caption{Average ranks on the 20 MTS datasets. WEASEL+MUSE and MLSTM-FCN are the most accurate.\label{fig:Average-Ranks-on}}
\end{figure}

Figure~\ref{fig:Average-Ranks-on} shows a critical difference diagram (as introduced in~\cite{demvsar2006statistical}) over the average ranks of the different MTSC methods. Classifiers with the lowest (best) ranks are to the right. The group of classifiers that are not significantly different in their rankings are connected by a bar. The critical difference (CD) length at the top represents statistically significant differences. 

MLSTM-FCN and WEASEL+MUSE show the lowest overall ranks and are in the group of best classifiers. These are also significantly better than the baseline DTWi. When compared to the plain WEASEL classifier, we can see that the MUSE extension to WEASEL also leads to significantly better ranks ($6.05$ vs $2.45$). This is a result of using feature identifiers and using derivatives.

Overall, WEASEL+MUSE has $12$ wins (or ties) on the MTS datasets (Table~\ref{tab:accuracies}), which is the  highest of all classifiers. With a mean of $93.5\%$ it shows a similar average accuracy as MLSTN-FCN with mean accuracy $92.1\%$.

In the next section we look into the differences between MLSTM-FCN and WEASEL+MUSE and identify the domains for which each classifier is best suited for.

\subsection{Domain-dependent strength or limitation}\label{subsec:accuracy_by_dataset_and_domain}

\begin{figure*}[t]
	\includegraphics[width=2\columnwidth]{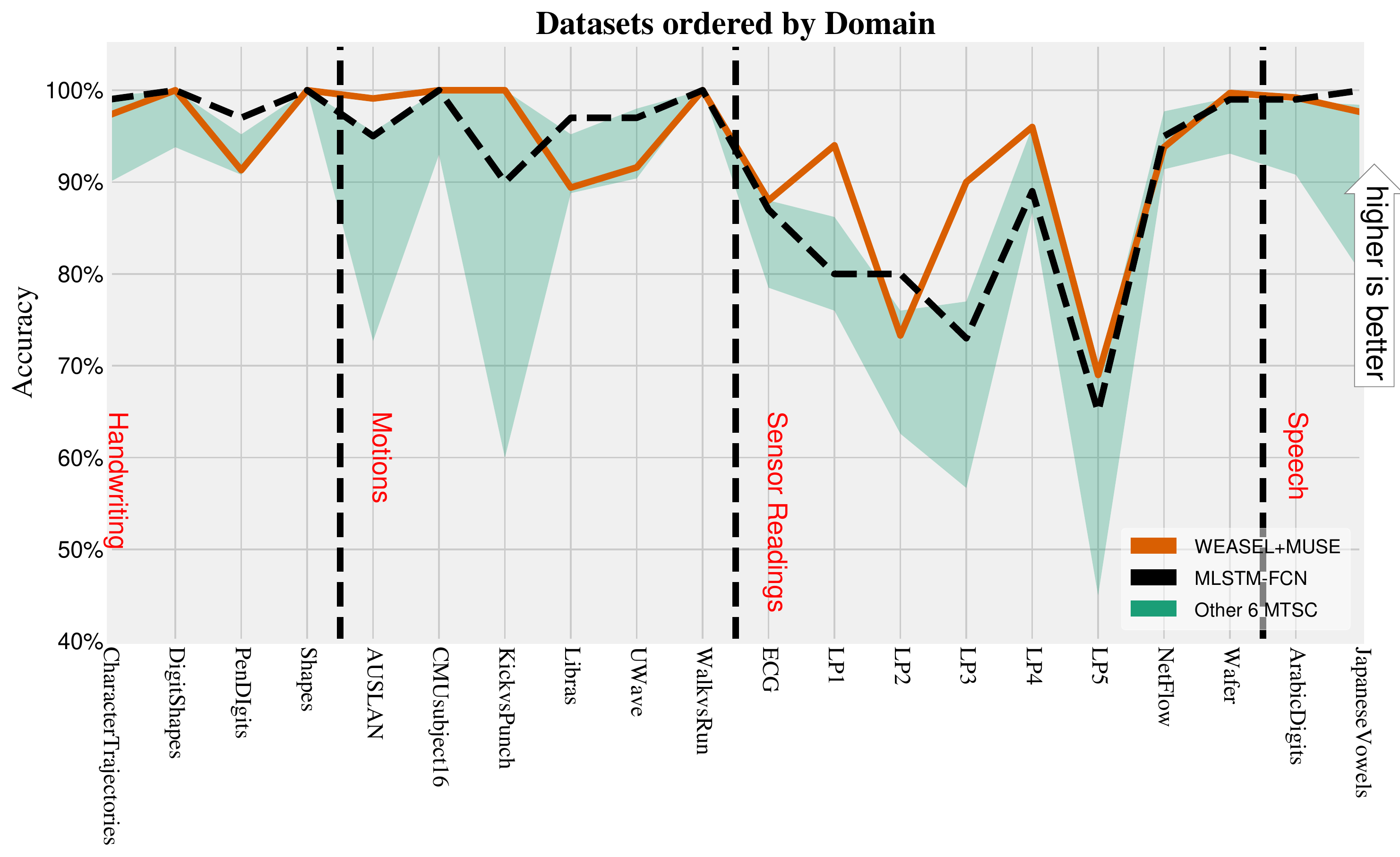}
	\caption{Classification accuracies on the $20$ MTS datasets for WEASEL+MUSE (orange), MLSTM-FCN (black) vs six state-of-the-art MTSC. The green area represents the classifiers' accuracies.\label{fig:Classification-accuracy-for}}
\end{figure*}

We studied the individual accuracy of each method on the 20 different MTS datasets, and grouped datasets by domain (Handwriting, Motion Sensors, Sensor Readings, Speech) to test if our method has a domain-dependent strength or limitation. 
Figure~\ref{fig:Classification-accuracy-for} shows the accuracies of WEASEL+MUSE (orange line), MLSTM-FCN (black line)  vs. the other six MTS classifiers (green area).

Overall, the performance of WEASEL+MUSE is very competitive for all datasets. The black line is mostly very close to the upper outline of the orange area, indicating that WEASEL+MUSE's performance is close to that of its best competitor in many cases. In total WEASEL+MUSE has $12$ out of $20$ possible wins (or ties). WEASEL+MUSE has the highest percentage of wins in the  groups of motion sensors, followed by speech and handwriting. 

WEASEL+MUSE and MLSTM-FCN perform similar on many dataset domain. WEASEL+MUSE performs best for sensor reading datasets and MLSTM-FCN performs best for motion and speech datasets. Sensor readings are the datasets with the least number of samples $N$ or features $n$ in the range of a few dozens. On the other hand, speech and motion datasets contain the most samples or features in the range of hundreds to thousands. 

This might indicate that WEASEL+MUSE performs well, even for small-sized datasets, whereas MLSTM-FCN seems to require larger training corpora to be most accurate. 
Furthermore, WEASEL+MUSE is based on the BOP model that compares signal based on the frequency of occurrence of subsequences rather than their absence or presence. Thus, signals with some repetition profit from using WEASEL+MUSE, such as ECG-signals.

\subsection{Effects of Gaussian Noise on Classification Accuracy}\label{subsec:noise}

\begin{figure*}[t]
	\begin{centering}
		\includegraphics[width=1\columnwidth]{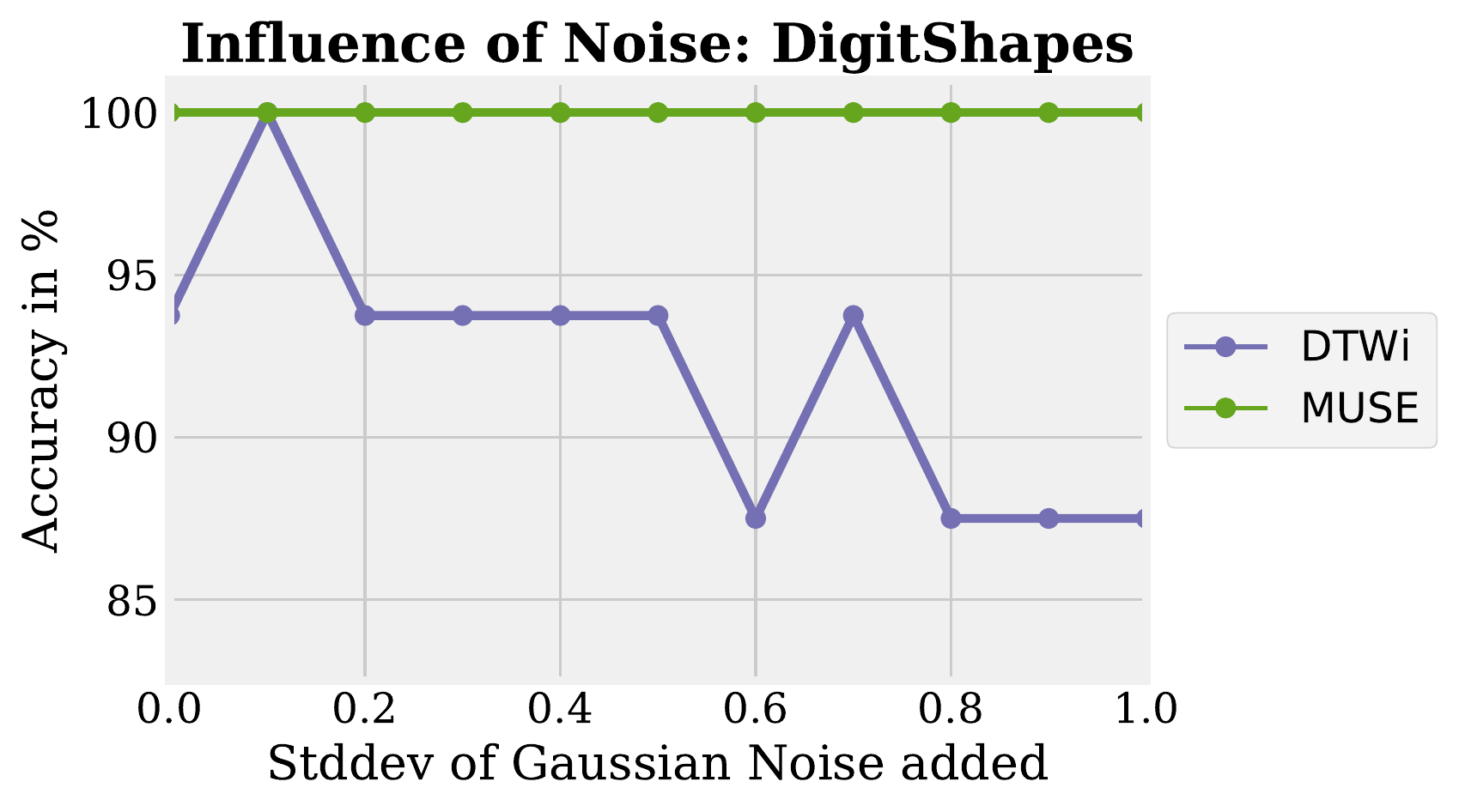}
		\includegraphics[width=1\columnwidth]{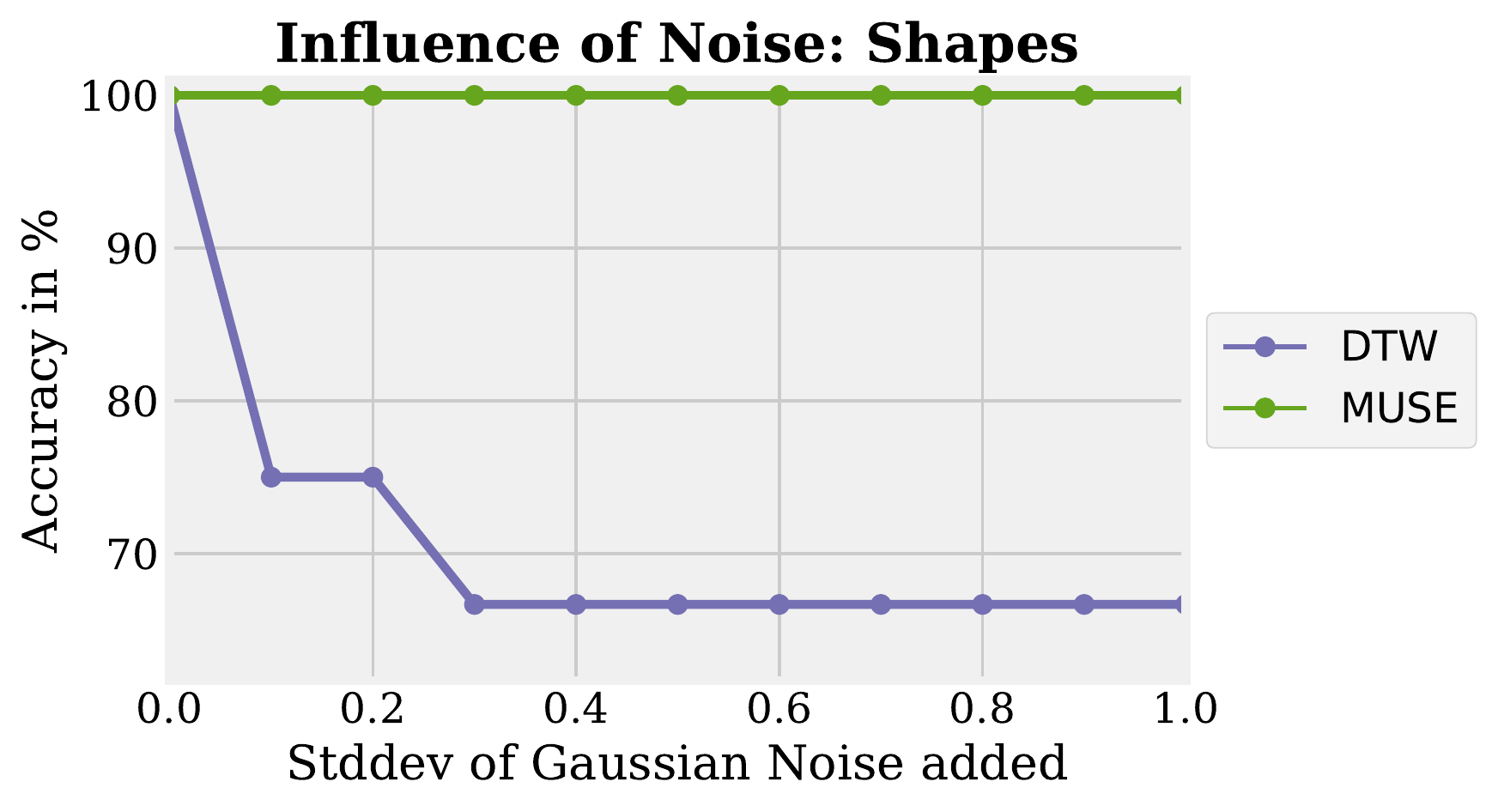}
	\end{centering}
	\caption{Effects of Gaussian noise on classification accuracy. With increasing levels of noise added, the accuracy of DTW drops and remains stable for WEASEL+MUSE. \label{fig:shapes_noise}}
\end{figure*}

WEASEL+MUSE applies the truncated Fourier Transform and discretization to generate features. This acts as a low-pass filter. To illustrate the relevance of noise to the classification task, we performed another experiment on the two multivariate synthetic datasets Shapes and DigitShapes.
 
First, all dimensions of each dataset were z-normalised to have a standard deviation (SD) of $1$. We then added an increasing Gaussian noise with a SD of $0$ to $1.0$ to each dimension, equal to noise levels of $0$\% to $100$\%. 

Figure \ref{fig:shapes_noise} shows the two classifiers DTWi and WEASEL+MUSE. For DTWi the classification accuracy drops by up to $30$ percentage points for increasing levels of noise. At the same time, WEASEL+MUSE was robust to Gaussian noise and its accuracy remains stable up to a noise level of $100\%$. 

%
%
%

\subsection{Relative Prediction Times}\label{subsec:influence}

\begin{table}[t]
	\begin{centering}
		\begin{tabular*}{1\columnwidth}{@{\extracolsep{\fill}}ccccc}
			& \multicolumn{2}{c}{Relative Time} & \multicolumn{2}{c}{Absolute Time in ms}\tabularnewline			
			& DTWi & MUSE & DTWi & MUSE\tabularnewline
			\hline 
			\hline 
			ArabicDigits & 206.3 & 1 & 10509055 & 50952\tabularnewline
			\hline 
			AUSLAN & 42.4 & 1 & 3040070 & 71737\tabularnewline
			\hline 
			CharTrajectories & 7.2 & 1 & 1153620 & 161104\tabularnewline
			\hline 
			CMU\_MOCAP & 0.4 & 1 & 162131 & 410387\tabularnewline
			\hline 
			DigitShapes & 16.7 & 1 & 2194 & 131\tabularnewline
			\hline 
			ECG & 2.1 & 1 & 4725 & 2228\tabularnewline
			\hline 
			Japanese Vowels & 34.1 & 1 & 15588 & 457\tabularnewline
			\hline 
			KickvsPunch & 0.1 & 1 & 31761 & 256406\tabularnewline
			\hline 
			LIBRAS & 9.7 & 1 & 3399 & 350\tabularnewline
			\hline 
			Robot Failure LP1 & 413.7 & 1 & 28961 & 70\tabularnewline
			\hline 
			Robot Failure LP2 & 1.2 & 1 & 53 & 43\tabularnewline
			\hline 
			Robot Failure LP3 & 1.4 & 1 & 51 & 36\tabularnewline
			\hline 
			Robot Failure LP4 & 10.1 & 1 & 689 & 68\tabularnewline
			\hline 
			Robot Failure LP5 & 20.4 & 1 & 1630 & 80\tabularnewline
			\hline 
			PenDigits & 45.6 & 1 & 21409 & 469\tabularnewline
			\hline 
			Shapes & 2.3 & 1 & 264 & 116\tabularnewline
			\hline 
			UWave & 4.5 & 1 & 5724667 & 1262706\tabularnewline
			\hline 
			Wafer & 8.8 & 1 & 704649 & 79945\tabularnewline
			\hline 
			WalkvsRun & 0.3 & 1 & 67350 & 242181\tabularnewline		
			\hline 
			\hline 
			{Average} & {43.3} & {1} & {1130119} & {133656}\tabularnewline
		\end{tabular*}
		\par\end{centering}
	\caption{Relative and absolute prediction times (lower is always better) of WEASEL+MUSE compared to DTWi.}
\end{table}

In addition to achieving state-of-the-art accuracy, WEASEL+MUSE is also competitive in terms of prediction times. In this experiment, we compare WEASEL+MUSE to DTWi. We could not perform a meaningful comparison to the other competitors, as we either do not have the source codes or the implementation is given in a different language (R, Matlab).

In general, 1-NN DTW has a computational complexity of $O(N n^{2})$ for TS of length $n$. For the implementation of DTWi we make use of the state-of-the-art cascading lower bounds from~\cite{rakthanmanon2012searching}. In this experiment, we measure CPU time to address parallel and single threaded codes. 
The DTWi prediction times is reported relative to that of WEASEL+MUSE, i.e., a number lower than 1 means that DTW is faster than WEASEL+MUSE. 1-NN DTW is a neatest neighbour classifier, so its prediction times directly depend on the size of the train dataset. For WEASEL+MUSE the length of the time series $n$ and the number of dimensions $m$ are most important

For all but three datasets WEASEL+MUSE is faster than DTWi. It is up to 400 times faster for the Robot Failure LP1 dataset and 200 times faster for ArabicDigits. On average it is 43 times faster than DTWi. There are three datasets for which DTW is faster: WalkvsRun, KickvsPunch and CMU-MOCAP. These are the datasets with the highest number of dimensions $m=62$. 
Thus, WEASEL+MUSE is not only significantly more accurate then DTWi but also orders of magnitude faster.

%
%

\subsection{Influence of Design Decisions on Accuracy}\label{subsec:influence}

\begin{figure}[h]
	\includegraphics[width=1\columnwidth]{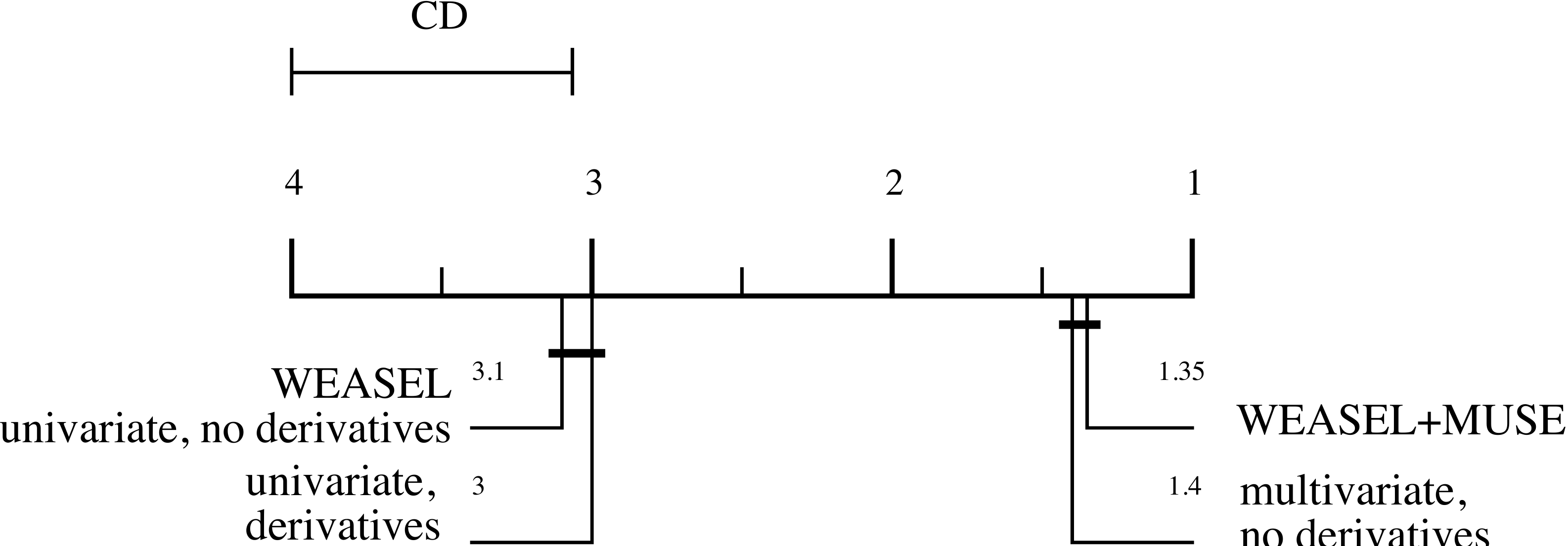}
	\caption{Impact of the design decisions of the WEASEL+MUSE classifier on accuracy.\label{fig:Impact-of-design}}
\end{figure}

We next look into the impact of several design decisions of the WEASEL+MUSE classifier. Figure~\ref{fig:Impact-of-design} shows the average ranks of the WEASEL+MUSE classifier on the 20 MTS datasets where each of the following extension is disabled or enabled: 

 \begin{enumerate}
	\item \emph{Multivariate vs Univariate} A key design decision of WEASEL+MUSE is to keep sensor ids for each word. The opposite is to treat all dimensions equally, i.e., concatenate all dimensionality information and treat the data like a univariate TS classification problem.
	\item \emph{Derivatives vs raw TS:} Following~\cite{wistuba2015ultra} and~\cite{baydogan2015learning}, we have added derivatives for each dimension to add trend information.
 \end{enumerate}

The \emph{univariate without derivatives} approach is in concept similar to WEASEL (without MUSE). There is a big gap between the multivariate and univariate models of WEASEL+MUSE. The univariate approaches are the least accurate, as the association of features to sensors is lost. The use of derivatives results in a slightly better score. Both extensions (multivariate and derivatives) combined improve ranks the most.

\subsection{Use Case: Motion Capture Data (Kinect)}\label{subsec:usecase}

This real world dataset was part of a challenge for gesture recognition~\cite{AALTD_CHALLENGE_16} and represents isolated gestures of different users captured by a Kinect camera system. The task was to predict the gestures performed by the users. 

The dataset consists of $180$ labelled train and $180$ test MTS. The labels on the test set were not revealed. A total of $8$ sensors were used to record x,y,z coordinates with a total of $51$ time instances, i.e., an MTS with $24$ streams of $51$ values each. The sensors are placed at the left/right hand, left/right elbow, left/right wrist and left/right thumb (see Figure~\ref{fig:model_mts_kinect} for an example gesture). 

WEASEL+MUSE (alias MWSL) scored $171$ correct predictions, which is equal to a test accuracy of $95\%$. The winning approach scored $173$ ($96.1\%$), based on an ensemble of random shapelets and domain specific feature extraction, and the SMTS~\cite{baydogan2015learning} classifier scored $172$ ($95.6\%$). 
This challenge underlined that WEASEL+MUSE is applicable out-of-the-box to real-world use cases and competitive with domain-specific tailored approaches.
Motion captured data is characterized by noisy data, that contains many superfluous information among dimensions. By design WEASEL+MUSE is able to deal with this kind of data effectively.

\section{Conclusion}

In this work, we have presented WEASEL+MUSE, a novel multivariate time series classification method following the bag-of-pattern approach and achieving highly competitive classification accuracies. The novelty of WEASEL+MUSE is its feature space engineering using statistical feature selection, derivatives, variable window lengths, bi-grams, and a symbolic representation for generating discriminative words. WEASEL+MUSE provides tolerance to noise (by use of the truncated Fourier transform), phase invariance, and superfluous data/dimensions. Thereby, WEASEL+MUSE assigns high weights to characteristic, local and global substructures along dimensions of a multivariate time series. In our evaluation on altogether 21 datasets, WEASEL+MUSE is consistently among the most accurate classifiers and outperforms state-of-the-art similarity measures or shapelet-based approaches. It performs well even for small-sized  datasets, where deep learning based approaches typically tend to perform poorly. When looking into application domains, it is best for sensor readings, followed by speech, motion and handwriting recognition tasks.

Future work could direct into different feature selection methods, benchmarking approaches based on train and prediction times, or use ensembling to build more powerful classifiers, which has been successfully used for univariate time series classification.

\bibliographystyle{ACM-Reference-Format}
\bibliography{muse}


\begin{thebibliography}{29}


\ifx \showCODEN    \undefined \def \showCODEN     #1{\unskip}     \fi
\ifx \showDOI      \undefined \def \showDOI       #1{#1}\fi
\ifx \showISBNx    \undefined \def \showISBNx     #1{\unskip}     \fi
\ifx \showISBNxiii \undefined \def \showISBNxiii  #1{\unskip}     \fi
\ifx \showISSN     \undefined \def \showISSN      #1{\unskip}     \fi
\ifx \showLCCN     \undefined \def \showLCCN      #1{\unskip}     \fi
\ifx \shownote     \undefined \def \shownote      #1{#1}          \fi
\ifx \showarticletitle \undefined \def \showarticletitle #1{#1}   \fi
\ifx \showURL      \undefined \def \showURL       {\relax}        \fi
\providecommand\bibfield[2]{#2}
\providecommand\bibinfo[2]{#2}
\providecommand\natexlab[1]{#1}
\providecommand\showeprint[2][]{arXiv:#2}

\bibitem[\protect\citeauthoryear{{AALTD Time Series Classification
  Challenge}}{{AALTD Time Series Classification Challenge}}{2016}]%
        {AALTD_CHALLENGE_16}
\bibfield{author}{\bibinfo{person}{{AALTD Time Series Classification
  Challenge}}.} \bibinfo{year}{2016}\natexlab{}.
\newblock \bibinfo{howpublished}{\url{https://aaltd16.irisa.fr/challenge/}}.
  (\bibinfo{year}{2016}).
\newblock


\bibitem[\protect\citeauthoryear{Bagnall, Lines, Bostrom, Large, and
  Keogh}{Bagnall et~al\mbox{.}}{2016}]%
        {bagnall2016great}
\bibfield{author}{\bibinfo{person}{Anthony Bagnall}, \bibinfo{person}{Jason
  Lines}, \bibinfo{person}{Aaron Bostrom}, \bibinfo{person}{James Large}, {and}
  \bibinfo{person}{Eamonn Keogh}.} \bibinfo{year}{2016}\natexlab{}.
\newblock \showarticletitle{{The Great Time Series Classification Bake Off: An
  Experimental Evaluation of Recently Proposed Algorithms. Extended Version}}.
\newblock \bibinfo{journal}{\emph{Data Mining and Knowledge Discovery}}
  (\bibinfo{year}{2016}), \bibinfo{pages}{1--55}.
\newblock


\bibitem[\protect\citeauthoryear{Baydogan and Runger}{Baydogan and
  Runger}{2015}]%
        {baydogan2015learning}
\bibfield{author}{\bibinfo{person}{Mustafa~Gokce Baydogan} {and}
  \bibinfo{person}{George Runger}.} \bibinfo{year}{2015}\natexlab{}.
\newblock \showarticletitle{Learning a symbolic representation for multivariate
  time series classification}.
\newblock \bibinfo{journal}{\emph{Data Mining and Knowledge Discovery}}
  \bibinfo{volume}{29}, \bibinfo{number}{2} (\bibinfo{year}{2015}),
  \bibinfo{pages}{400--422}.
\newblock


\bibitem[\protect\citeauthoryear{Baydogan and Runger}{Baydogan and
  Runger}{2016}]%
        {baydogan2016time}
\bibfield{author}{\bibinfo{person}{Mustafa~Gokce Baydogan} {and}
  \bibinfo{person}{George Runger}.} \bibinfo{year}{2016}\natexlab{}.
\newblock \showarticletitle{Time series representation and similarity based on
  local autopatterns}.
\newblock \bibinfo{journal}{\emph{Data Mining and Knowledge Discovery}}
  \bibinfo{volume}{30}, \bibinfo{number}{2} (\bibinfo{year}{2016}),
  \bibinfo{pages}{476--509}.
\newblock


\bibitem[\protect\citeauthoryear{Cuturi and Doucet}{Cuturi and Doucet}{2011}]%
        {cuturi2011autoregressive}
\bibfield{author}{\bibinfo{person}{Marco Cuturi} {and} \bibinfo{person}{Arnaud
  Doucet}.} \bibinfo{year}{2011}\natexlab{}.
\newblock \showarticletitle{Autoregressive kernels for time series}.
\newblock \bibinfo{journal}{\emph{arXiv preprint arXiv:1101.0673}}
  (\bibinfo{year}{2011}).
\newblock


\bibitem[\protect\citeauthoryear{Dem{\v{s}}ar}{Dem{\v{s}}ar}{2006}]%
        {demvsar2006statistical}
\bibfield{author}{\bibinfo{person}{Janez Dem{\v{s}}ar}.}
  \bibinfo{year}{2006}\natexlab{}.
\newblock \showarticletitle{{Statistical comparisons of classifiers over
  multiple data sets}}.
\newblock \bibinfo{journal}{\emph{{The Journal of Machine Learning Research}}}
  \bibinfo{volume}{7} (\bibinfo{year}{2006}), \bibinfo{pages}{1--30}.
\newblock


\bibitem[\protect\citeauthoryear{Esling and Agon}{Esling and Agon}{2012}]%
        {esling2012time}
\bibfield{author}{\bibinfo{person}{Philippe Esling} {and}
  \bibinfo{person}{Carlos Agon}.} \bibinfo{year}{2012}\natexlab{}.
\newblock \showarticletitle{{Time-series data mining}}.
\newblock \bibinfo{journal}{\emph{{ACM Computing Surveys}}}
  \bibinfo{volume}{45}, \bibinfo{number}{1} (\bibinfo{year}{2012}),
  \bibinfo{pages}{12:1--12:34}.
\newblock


\bibitem[\protect\citeauthoryear{Fan, Chang, Hsieh, Wang, and Lin}{Fan
  et~al\mbox{.}}{2008}]%
        {fan2008liblinear}
\bibfield{author}{\bibinfo{person}{Rong-En Fan}, \bibinfo{person}{Kai-Wei
  Chang}, \bibinfo{person}{Cho-Jui Hsieh}, \bibinfo{person}{Xiang-Rui Wang},
  {and} \bibinfo{person}{Chih-Jen Lin}.} \bibinfo{year}{2008}\natexlab{}.
\newblock \showarticletitle{LIBLINEAR: A library for large linear
  classification}.
\newblock \bibinfo{journal}{\emph{The Journal of Machine Learning Research}}
  \bibinfo{volume}{9} (\bibinfo{year}{2008}), \bibinfo{pages}{1871--1874}.
\newblock


\bibitem[\protect\citeauthoryear{Hobbs, Jitprapaikulsarn, Konda, Chankong,
  Loparo, and Maratukulam}{Hobbs et~al\mbox{.}}{1999}]%
        {hobbs1999analysis}
\bibfield{author}{\bibinfo{person}{Benjamin~F Hobbs}, \bibinfo{person}{Suradet
  Jitprapaikulsarn}, \bibinfo{person}{Sreenivas Konda}, \bibinfo{person}{Vira
  Chankong}, \bibinfo{person}{Kenneth~A Loparo}, {and}
  \bibinfo{person}{Dominic~J Maratukulam}.} \bibinfo{year}{1999}\natexlab{}.
\newblock \showarticletitle{Analysis of the value for unit commitment of
  improved load forecasts}.
\newblock \bibinfo{journal}{\emph{IEEE Transactions on Power Systems}}
  \bibinfo{volume}{14}, \bibinfo{number}{4} (\bibinfo{year}{1999}),
  \bibinfo{pages}{1342--1348}.
\newblock


\bibitem[\protect\citeauthoryear{Jerzak and Ziekow}{Jerzak and Ziekow}{2014}]%
        {jerzak2014debs}
\bibfield{author}{\bibinfo{person}{Zbigniew Jerzak} {and}
  \bibinfo{person}{Holger Ziekow}.} \bibinfo{year}{2014}\natexlab{}.
\newblock \showarticletitle{{The DEBS 2014 Grand Challenge}}. In
  \bibinfo{booktitle}{\emph{{Proceedings of the 2014 ACM International
  Conference on Distributed Event-based Systems}}}. ACM,
  \bibinfo{pages}{266--269}.
\newblock


\bibitem[\protect\citeauthoryear{Karim, Majumdar, Darabi, and Harford}{Karim
  et~al\mbox{.}}{2018}]%
        {karim2018multivariate}
\bibfield{author}{\bibinfo{person}{Fazle Karim}, \bibinfo{person}{Somshubra
  Majumdar}, \bibinfo{person}{Houshang Darabi}, {and} \bibinfo{person}{Samuel
  Harford}.} \bibinfo{year}{2018}\natexlab{}.
\newblock \showarticletitle{Multivariate LSTM-FCNs for Time Series
  Classification}.
\newblock \bibinfo{journal}{\emph{arXiv preprint arXiv:1801.04503}}
  (\bibinfo{year}{2018}).
\newblock


\bibitem[\protect\citeauthoryear{Karlsson, Papapetrou, and
  Bostr{\"o}m}{Karlsson et~al\mbox{.}}{2016}]%
        {karlsson2016generalized}
\bibfield{author}{\bibinfo{person}{Isak Karlsson}, \bibinfo{person}{Panagiotis
  Papapetrou}, {and} \bibinfo{person}{Henrik Bostr{\"o}m}.}
  \bibinfo{year}{2016}\natexlab{}.
\newblock \showarticletitle{Generalized random shapelet forests}.
\newblock \bibinfo{journal}{\emph{Data mining and knowledge discovery}}
  \bibinfo{volume}{30}, \bibinfo{number}{5} (\bibinfo{year}{2016}),
  \bibinfo{pages}{1053--1085}.
\newblock


\bibitem[\protect\citeauthoryear{Lin, Keogh, Wei, and Lonardi}{Lin
  et~al\mbox{.}}{2007}]%
        {Lin2007}
\bibfield{author}{\bibinfo{person}{Jessica Lin}, \bibinfo{person}{Eamonn~J.
  Keogh}, \bibinfo{person}{Li Wei}, {and} \bibinfo{person}{Stefano Lonardi}.}
  \bibinfo{year}{2007}\natexlab{}.
\newblock \showarticletitle{{Experiencing SAX: a novel symbolic representation
  of time series}}.
\newblock \bibinfo{journal}{\emph{{Data Mining and knowledge discovery}}}
  \bibinfo{volume}{15}, \bibinfo{number}{2} (\bibinfo{year}{2007}),
  \bibinfo{pages}{107--144}.
\newblock


\bibitem[\protect\citeauthoryear{Lin, Khade, and Li}{Lin et~al\mbox{.}}{2012}]%
        {0001KL12}
\bibfield{author}{\bibinfo{person}{Jessica Lin}, \bibinfo{person}{Rohan Khade},
  {and} \bibinfo{person}{Yuan Li}.} \bibinfo{year}{2012}\natexlab{}.
\newblock \showarticletitle{{Rotation-invariant similarity in time series using
  bag-of-patterns representation}}.
\newblock \bibinfo{journal}{\emph{{Journal of Intelligent Information
  Systems}}} \bibinfo{volume}{39}, \bibinfo{number}{2} (\bibinfo{year}{2012}),
  \bibinfo{pages}{287--315}.
\newblock


\bibitem[\protect\citeauthoryear{{Mustafa Gokce Baydogan}}{{Mustafa Gokce
  Baydogan}}{2017}]%
        {MultivariateTimeSeriesClassification}
\bibfield{author}{\bibinfo{person}{{Mustafa Gokce Baydogan}}.}
  \bibinfo{year}{2017}\natexlab{}.
\newblock \bibinfo{title}{{Multivariate Time Series Classification Datasets}}.
\newblock \bibinfo{howpublished}{\url{http://www.mustafabaydogan.com}}.
  (\bibinfo{year}{2017}).
\newblock


\bibitem[\protect\citeauthoryear{Mutschler, Ziekow, and Jerzak}{Mutschler
  et~al\mbox{.}}{2013}]%
        {mutschler2013debs}
\bibfield{author}{\bibinfo{person}{Christopher Mutschler},
  \bibinfo{person}{Holger Ziekow}, {and} \bibinfo{person}{Zbigniew Jerzak}.}
  \bibinfo{year}{2013}\natexlab{}.
\newblock \showarticletitle{{The DEBS 2013 grand challenge}}. In
  \bibinfo{booktitle}{\emph{{Proceedings of the 2013 ACM International
  Conference on Distributed Event-based Systems}}}. ACM,
  \bibinfo{pages}{289--294}.
\newblock


\bibitem[\protect\citeauthoryear{Ng}{Ng}{2004}]%
        {ng2004feature}
\bibfield{author}{\bibinfo{person}{Andrew~Y Ng}.}
  \bibinfo{year}{2004}\natexlab{}.
\newblock \showarticletitle{Feature selection, L 1 vs. L 2 regularization, and
  rotational invariance}. In \bibinfo{booktitle}{\emph{{Proceedings of the 2004
  ACM International Conference on Machine Learning}}}. ACM,
  \bibinfo{pages}{78}.
\newblock


\bibitem[\protect\citeauthoryear{Rakthanmanon, Campana, Mueen, Batista,
  Westover, Zhu, Zakaria, and Keogh}{Rakthanmanon et~al\mbox{.}}{2012}]%
        {rakthanmanon2012searching}
\bibfield{author}{\bibinfo{person}{Thanawin Rakthanmanon},
  \bibinfo{person}{Bilson Campana}, \bibinfo{person}{Abdullah Mueen},
  \bibinfo{person}{Gustavo Batista}, \bibinfo{person}{Brandon Westover},
  \bibinfo{person}{Qiang Zhu}, \bibinfo{person}{Jesin Zakaria}, {and}
  \bibinfo{person}{Eamonn Keogh}.} \bibinfo{year}{2012}\natexlab{}.
\newblock \showarticletitle{{Searching and mining trillions of time series
  subsequences under dynamic time warping}}. In
  \bibinfo{booktitle}{\emph{{Proceedings of the 2012 ACM SIGKDD International
  Conference on Knowledge Discovery and Data Mining}}}. ACM,
  \bibinfo{pages}{262--270}.
\newblock


\bibitem[\protect\citeauthoryear{Sch{\"a}fer}{Sch{\"a}fer}{2015a}]%
        {schafer2015scalable2}
\bibfield{author}{\bibinfo{person}{Patrick Sch{\"a}fer}.}
  \bibinfo{year}{2015}\natexlab{a}.
\newblock \showarticletitle{Scalable time series classification}.
\newblock \bibinfo{journal}{\emph{Data Mining and Knowledge Discovery}}
  (\bibinfo{year}{2015}), \bibinfo{pages}{1--26}.
\newblock


\bibitem[\protect\citeauthoryear{Sch{\"a}fer}{Sch{\"a}fer}{2015b}]%
        {schafer2014boss}
\bibfield{author}{\bibinfo{person}{Patrick Sch{\"a}fer}.}
  \bibinfo{year}{2015}\natexlab{b}.
\newblock \showarticletitle{{The BOSS is concerned with time series
  classification in the presence of noise}}.
\newblock \bibinfo{journal}{\emph{{Data Mining and Knowledge Discovery}}}
  \bibinfo{volume}{29}, \bibinfo{number}{6} (\bibinfo{year}{2015}),
  \bibinfo{pages}{1505--1530}.
\newblock


\bibitem[\protect\citeauthoryear{Sch\"{a}fer and H\"{o}gqvist}{Sch\"{a}fer and
  H\"{o}gqvist}{2012}]%
        {SchaferH12}
\bibfield{author}{\bibinfo{person}{Patrick Sch\"{a}fer} {and}
  \bibinfo{person}{Mikael H\"{o}gqvist}.} \bibinfo{year}{2012}\natexlab{}.
\newblock \showarticletitle{{SFA: a symbolic fourier approximation and index
  for similarity search in high dimensional datasets}}. In
  \bibinfo{booktitle}{\emph{{Proceedings of the 2012 International Conference
  on Extending Database Technology}}}. ACM, \bibinfo{pages}{516--527}.
\newblock


\bibitem[\protect\citeauthoryear{Sch{\"{a}}fer and Leser}{Sch{\"{a}}fer and
  Leser}{2017}]%
        {benchmarkingSchaefer17}
\bibfield{author}{\bibinfo{person}{Patrick Sch{\"{a}}fer} {and}
  \bibinfo{person}{Ulf Leser}.} \bibinfo{year}{2017}\natexlab{}.
\newblock \showarticletitle{Benchmarking Univariate Time Series Classifiers}.
  In \bibinfo{booktitle}{\emph{BTW 2017}}. \bibinfo{pages}{289--298}.
\newblock


\bibitem[\protect\citeauthoryear{Sch{\"a}fer and Leser}{Sch{\"a}fer and
  Leser}{2017}]%
        {schaefer2017weasel}
\bibfield{author}{\bibinfo{person}{Patrick Sch{\"a}fer} {and}
  \bibinfo{person}{Ulf Leser}.} \bibinfo{year}{2017}\natexlab{}.
\newblock \showarticletitle{{Fast and Accurate Time Series Classification with
  WEASEL}}.
\newblock \bibinfo{journal}{\emph{{Proceedings of the 2017 {ACM} on Conference
  on Information and Knowledge Management}}} (\bibinfo{year}{2017}),
  \bibinfo{pages}{637--646}.
\newblock


\bibitem[\protect\citeauthoryear{{The Value of Wind Power Forecasting}}{{The
  Value of Wind Power Forecasting}}{2016}]%
        {WindPower}
\bibfield{author}{\bibinfo{person}{{The Value of Wind Power Forecasting}}.}
  \bibinfo{year}{2016}\natexlab{}.
\newblock
  \bibinfo{howpublished}{\url{http://www.nrel.gov/docs/fy11osti/50814.pdf}}.
  (\bibinfo{year}{2016}).
\newblock


\bibitem[\protect\citeauthoryear{Tuncel and Baydogan}{Tuncel and
  Baydogan}{2018}]%
        {tuncel2018autoregressive}
\bibfield{author}{\bibinfo{person}{Kerem~Sinan Tuncel} {and}
  \bibinfo{person}{Mustafa~Gokce Baydogan}.} \bibinfo{year}{2018}\natexlab{}.
\newblock \showarticletitle{Autoregressive forests for multivariate time series
  modeling}.
\newblock \bibinfo{journal}{\emph{Pattern Recognition}}  \bibinfo{volume}{73}
  (\bibinfo{year}{2018}), \bibinfo{pages}{202--215}.
\newblock


\bibitem[\protect\citeauthoryear{{WEASEL+MUSE Classifier Source Code and Raw
  Results}}{{WEASEL+MUSE Classifier Source Code and Raw Results}}{2017}]%
        {MUSEWebPage}
\bibfield{author}{\bibinfo{person}{{WEASEL+MUSE Classifier Source Code and Raw
  Results}}.} \bibinfo{year}{2017}\natexlab{}.
\newblock
  \bibinfo{howpublished}{\url{https://www2.informatik.hu-berlin.de/~schaefpa/muse/}}.
    (\bibinfo{year}{2017}).
\newblock


\bibitem[\protect\citeauthoryear{Wistuba, Grabocka, and Schmidt-Thieme}{Wistuba
  et~al\mbox{.}}{2015}]%
        {wistuba2015ultra}
\bibfield{author}{\bibinfo{person}{Martin Wistuba}, \bibinfo{person}{Josif
  Grabocka}, {and} \bibinfo{person}{Lars Schmidt-Thieme}.}
  \bibinfo{year}{2015}\natexlab{}.
\newblock \showarticletitle{Ultra-fast shapelets for time series
  classification}.
\newblock \bibinfo{journal}{\emph{arXiv preprint arXiv:1503.05018}}
  (\bibinfo{year}{2015}).
\newblock


\bibitem[\protect\citeauthoryear{{Y Chen, E Keogh, B Hu, N Begum, A Bagnall, A
  Mueen and G Batista }}{{Y Chen, E Keogh, B Hu, N Begum, A Bagnall, A Mueen
  and G Batista }}{2015}]%
        {UCRClassification}
\bibfield{author}{\bibinfo{person}{{Y Chen, E Keogh, B Hu, N Begum, A Bagnall,
  A Mueen and G Batista }}.} \bibinfo{year}{2015}\natexlab{}.
\newblock \bibinfo{title}{{The UCR Time Series Classification Archive}}.
\newblock
  \bibinfo{howpublished}{\url{http://www.cs.ucr.edu/~eamonn/time_series_data}}.
    (\bibinfo{year}{2015}).
\newblock


\bibitem[\protect\citeauthoryear{Ye and Keogh}{Ye and Keogh}{2009}]%
        {YeK09}
\bibfield{author}{\bibinfo{person}{Lexiang Ye} {and} \bibinfo{person}{Eamonn~J.
  Keogh}.} \bibinfo{year}{2009}\natexlab{}.
\newblock \showarticletitle{{Time series shapelets: a new primitive for data
  mining}}. In \bibinfo{booktitle}{\emph{{Proceedings of the 2009 ACM SIGKDD
  International Conference on Knowledge Discovery and Data Mining}}}.
  \bibinfo{publisher}{ACM}.
\newblock


\end{thebibliography}

\end{document}